\documentclass[12pt,onecolumn,letterpaper]{article}

\usepackage{iccv}
\usepackage{times}
\usepackage{epsfig}
\usepackage{graphicx}
\usepackage{amsmath}
\usepackage{amssymb}

\usepackage{multirow}
\usepackage{soul}
\usepackage{subcaption}
\usepackage{enumitem}

\usepackage{bbm}
\usepackage{colortbl}
\usepackage{comment}
\usepackage{appendix}
\usepackage{etoolbox}
\usepackage{lipsum}
\usepackage[dvipsnames]{xcolor}
\newcommand{\cmark}{\ding{51}}
\newcommand{\xmark}{\ding{55}}
\usepackage{pifont}
\usepackage{tabularx}
\usepackage{booktabs}

\definecolor{mygray}{gray}{0.6}

\newcommand\blfootnote[1]{%
  \begingroup
  \renewcommand\thefootnote{}\thanks{#1}%
  \endgroup
}

\usepackage[pagebackref=true,breaklinks=true,letterpaper=true,colorlinks,bookmarks=false]{hyperref}

\iccvfinalcopy 

\def\iccvPaperID{1259} 

\ificcvfinal\fi

\begin{document}

\title{Box2Mask: Box-supervised Instance Segmentation via \\ Level-set Evolution}

\author{{\normalsize Wentong Li$^1$, Wenyu Liu$^1$, Jianke Zhu$^{1,3}$, Miaomiao Cui$^2$,} \\ {\normalsize Risheng Yu$^3$,  Xiansheng Hua$^2$, Lei Zhang$^4$} \\ 
{\normalsize $^1$Zhejiang University \ \ \ $^2$DAMO Academy, Alibaba Group} \\
{\normalsize $^3$The Second Affiliated Hospital Zhejiang University School of Medicine} \\
{\normalsize $^4$Dept. of Computing, The Hong Kong Polytechnic University} \\
{\tt\small \{liwentong, liuwenyu.lwy, jkzhu, risheng-yu\}@zju.edu.cn} \\
{\tt\small miaomiao.cmm@alibaba-inc.com, xshua@outlook.com, cslzhang@comp.polyu.edu.hk}
\blfootnote{Corresponding author is Jianke Zhu}
}

\maketitle
\ificcvfinal\thispagestyle{empty}\fi

\begin{abstract}
      In contrast to fully supervised methods using pixel-wise mask labels, box-supervised instance segmentation takes advantage of simple box annotations, which has recently attracted increasing research attention. This paper presents a novel single-shot instance segmentation approach, namely Box2Mask, which integrates the classical level-set evolution model into deep neural network learning to achieve accurate mask prediction with only bounding box supervision.
      Specifically, both the input image and its deep features are employed to evolve the level-set curves implicitly, and a local consistency module based on a pixel affinity kernel is used to mine the local context and spatial relations.
      Two types of single-stage frameworks, \textit{i.e.,} CNN-based and transformer-based frameworks, are developed to empower the level-set evolution for box-supervised instance segmentation, and each framework consists of three essential components: instance-aware decoder, box-level matching assignment and level-set evolution. By minimizing the level-set energy function, the mask map of each instance can be iteratively optimized within its bounding box annotation. The experimental results on five challenging testbeds, covering general scenes, remote sensing, medical and scene text images, demonstrate the outstanding performance of our proposed Box2Mask approach for box-supervised instance segmentation. In particular, with the Swin-Transformer large backbone, our Box2Mask obtains 42.4\% mask AP on COCO, which is on par with the recently developed fully mask-supervised methods. The code is available at:  \href{https://github.com/LiWentomng/boxlevelset}{https://github.com/LiWentomng/boxlevelset}.
\end{abstract}

\section{Introduction}
\label{intro}
Instance segmentation aims to obtain the pixel-wise mask labels of the interested objects, which plays an important role in applications such as autonomous driving, robotic manipulation, image editing, cell segmentation, etc. Benefiting from the strong learning capacity of advanced CNN~\cite{he2016deep, densenet_cvpr2017, liu2022convnet} and transformer~\cite{vaswani2017attention, vit_iclr2020, liu2021swin} architectures, instance segmentation has achieved remarkable progresses in recent years.
However, many of the existing instance segmentation models~\cite{eccv2020boundary,cvpr2019mask,wang2020solov2,eccv2020_condinst,iccv2017maskrcnn, ICCV2021_queryinst, cvpr2022mask2former} are trained in a fully supervised manner, which heavily depend on the pixel-wise instance mask annotations and incur expensive and tedious labeling costs.

\begin{figure*}[t]
	\centering
	\includegraphics[width=0.98\linewidth]{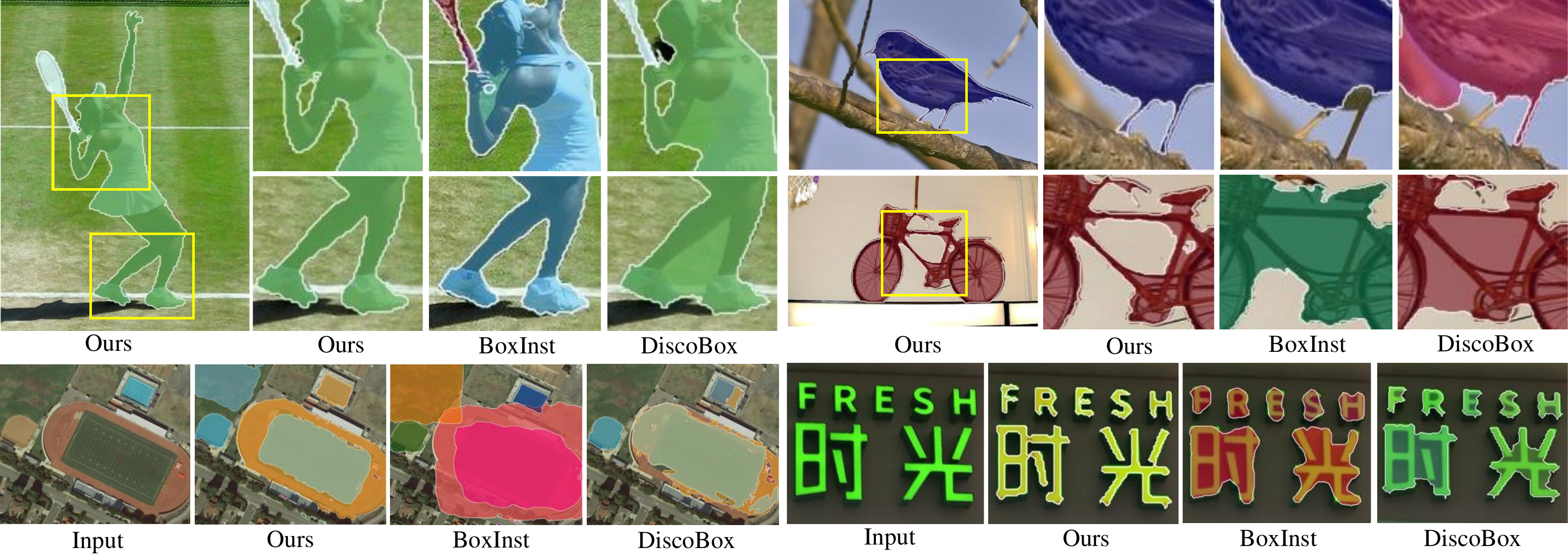}
	\caption{Qualitative comparison between our proposed Box2Mask against recent pairwise affinity modeling methods, \textit{i.e.,} BoxInst~\cite{cvpr2021_boxinst} and DiscoBox~\cite{iccv2021discobox}, on general scenes, remote sensing and scene text images. One can see that our method preserves finer details with accurate boundaries on various scenarios. All models are trained with only box annotations. Best viewed on screen.}
	\label{fig:introduction}
\end{figure*}

To alleviate the above problem, box-supervised instance segmentation has been proposed, which employs the simple and label-efficient box annotations rather than the pixel-wise mask labels. On average, it costs about 79.2 seconds to generate the polygon-based mask of an object on COCO~\cite{lin2014microsoft}, whereas it costs only 7 seconds~\cite{papadopoulos2017extreme} to annotate the bounding box of an object. 
Box annotation makes instance segmentation more accessible for new categories or scene types, which has recently attracted a lot of research attentions~\cite{eccv2020box2seg,nips2019-bbtp,cvpr2021_boxinst,cvpr2021bbam,cvpr2021boxcaseg,iccv2021discobox}. To enable pixel-wise supervision with box annotation, some methods~\cite{cvpr2021bbam,cvpr2021boxcaseg} have been developed by employing extra auxiliary salient data~\cite{cvpr2021boxcaseg} or post-processing methods like MCG~\cite{tpmai2017mcg} and CRF~\cite{krahenbuhl2011efficient} to obtain pseudo labels. However, these methods involve multiple separate steps, complicating the training pipeline and introducing many hyper-parameters to tune.
Several recent approaches~\cite{nips2019-bbtp,cvpr2021_boxinst} suggest a unified framework using the pairwise affinity modeling, e.g., neighbouring pixel pairs~\cite{nips2019-bbtp} and colour pairs~\cite{cvpr2021_boxinst}, enabling an end-to-end training of the instance segmentation network.
DiscoBox~\cite{iccv2021discobox} further incorporates intra-image and cross-image pairwise potentials to model the pixel affinities, and builds an independent teacher network to generate pseudo-labels as the supervision of student network. 
These methods define the pairwise affinity relationship on partial or all neighbouring pixel pairs, which oversimplifies the assumption that similar pixel or colour pairs are encouraged to share the same label. They are susceptible to objects with similar appearance or complicated background, resulting in inferior instance segmentation performance.

In this work, we explore more robust affinity modeling methods for effective box-supervised instance segmentation by using the classical level-set model~\cite{levelset1995a, osher1988fronts, tip2001_active_contour}, which employs an energy function to represent implicitly the object boundary curves. The level-set based energy function can exploit rich context information, including pixel intensity, color, appearance and shape, and has brought promising image segmentation results~\cite{eccv2020levelset, yuan2020deep, cvpr2019deeplevelsetevolution, CVPR2017hudeep}. However, these methods perform level-set evolution in a fully mask-supervised manner, where the network is trained to predict the object boundaries with pixel-wise supervision. Different from these methods, in this work we aim to supervise the training of level-set evolution with only bounding box annotations.  
Specifically, we propose a novel box-supervised instance segmentation approach, namely \textit{Box2Mask}, which delicately integrates the classical level-set model~\cite{osher1988fronts,tip2001_active_contour} with deep neural networks to iteratively learn a series of level-set functions for implicit curve evolution.
The classical continuous Chan-Vese energy functional~\cite{tip2001_active_contour} is used in our approach. 
We make use of both low-level and high-level features to robustly evolve the level-set curves towards the object's boundary. 
At each evolution step, the level-set is automatically initialized by a simple box projection function, which provides the rough estimation of the target boundary.
A local consistency module is developed based on an affinity kernel function, which mines the local context and spatial relations, to ensure the level-set evolution with local affinity consistency.

We present two types of single-stage frameworks to empower the level-set evolution,  
a CNN-based framework and a transformer-based framework. In addition to the level-set evolution part, each framework consists of another two essential components, e.g., instance-aware decoder (IAD) and box-level matching assignment, which are  equipped with different techniques. The IAD learns to embed the instance-wise characteristics, which generates dynamically the full-image instance-aware mask map as the level-set prediction conditioned on the input target instance. The  box-based matching assignment learns to assign the high-quality mask map samples as the positives with the guidance of ground truth bounding boxes.
By minimizing the fully differentiable level-set energy function, the mask map of each instance can be iteratively optimized within its corresponding bounding box annotation.

The preliminary results of our work have been reported in our conference paper~\cite{li2022box}. In this extended journal version, we first extend our method from the CNN-based framework to the transformer-based framework. To this end, we make use of the transformer decoder to embed instance-wise characteristics for dynamic kernel learning, and introduce a box-level bipartite matching scheme for label assignment. In addition, to alleviate the region-based intensity inhomogeneity of level-set evolution, we develop a local consistency module based on an affinity kernel function, which further mines the pixel similarities and spatial relationships within the neighborhood. 

Extensive experiments are conducted on five challenging testbeds for instance segmentation under various scenarios, including general scenes (\textit{i.e.} COCO~\cite{lin2014microsoft} and Pascal VOC~\cite{pascalvoc2010}), remote sensing, medical  and scene text images. The leading quantitative and qualitative results demonstrate the effectiveness of our proposed Box2Mask method. Specifically, it significantly improves the previous state-of-the-art $38.3\%$ AP to $43.2\%$ AP on Pascal VOC, and $33.4\%$ AP to $38.3\%$ AP on COCO with ResNet-101 backbone~\cite{he2016deep}.
It performs even better than some typical fully mask-supervised methods with the same backbone, e.g., Mask R-CNN~\cite{iccv2017maskrcnn}, SOLO~\cite{wang2020solo} and PolarMask~\cite{cvpr_2020polarmask}. With the stronger Swin-Transformer large (Swin-L) backbone~\cite{liu2021swin}, our Box2Mask can obtain \textbf{42.4\%} mask AP on COCO, which is on par with the recently well-established fully mask-supervised methods.
Some visual comparisons are shown in Fig.~\ref{fig:introduction}. One can see that the mask predictions of our method are generally of higher quality and with finer details than the recently developed BoxInst~\cite{cvpr2021_boxinst} and DiscoBox~\cite{iccv2021discobox} methods.

\section{Related Work}
Our work is related to fully-supervised instance segmentation methods, box-supervised instance segmentation methods, and level-set based segmentation methods.

\subsection{Fully-supervised Instance Segmentation}
Instance segmentation is a fundamental yet challenging computer vision task, which aims to generate a pixel-level mask with a category label for each individual instance of interest in an image. 
Most of the existing works are fully supervised approaches with pixel-wise mask supervision, which can be roughly divided into three categories. 

The first category is the Mask R-CNN family~\cite{iccv2017maskrcnn,cvpr2019cascadercnn, cvpr2020pointrend,eccv2020boundary,zhang2021refinemask}, which performs segmentation on the regions (\textit{e.g.}, ROIs) extracted from the detection results. The second category includes ROI-free approaches~\cite{xie2021polarmask++, eccv2020_condinst, wang2020solov2, nips2021knet}, which directly segments each instance in a fully convolutional manner without resorting to the detection regions.  
The above mentioned instance segmentation approaches
prevail with the convolutional neural network (CNN)-based framework, until recently transformer-based methods~\cite{eccv2020-detr, ICCV2021_queryinst, nips2021maskformer, cvpr2022mask2former, arxiv2022maskdino} have achieved remarkable progress.
These transformer based methods employ a set matching mechanism~\cite{eccv2020-detr}, and learn object queries to achieve the goal of instance segmentation.
Though being able to segment objects with accurate boundaries, these methods rely on the expensive and laborious pixel-wise mask annotations, which limits their deployments for new categories or scene types in real-world applications. The weakly supervised methods with label-efficient annotations have received increasing attention~\cite{tpami2020leveraging, papadopoulos2017extreme, TPAMI2021affinity}. In our work, we show that by using the simple bounding box obtained by some object detectors as the weak supervision, accurate instance segmentation results can still be achieved. 

\subsection{Box-supervised Instance Segmentation}
Box-supervised instance segmentation employs the simple bounding box annotations to obtain an accurate pixel-level mask prediction. Khoreva~\textit{et al.}~\cite{cvpr2017SDI} proposed to predict the mask with the box annotations under the deep learning framework, which heavily depends on the region proposals generated by the unsupervised segmentation methods like GrabCut~\cite{TOG2004grabcut} and MCG~\cite{tpmai2017mcg}. Based on the Mask R-CNN~\cite{iccv2017maskrcnn}, Hsu~\textit{et al.}~\cite{nips2019-bbtp} formulated the box-supervised instance segmentation problem as a multiple instance learning (MIL) problem by making use of the neighbouring pixel-pairwise affinity regularization. BoxInst~\cite{cvpr2021_boxinst} is proposed to use the color-pairwise affinity with box constraint under the efficient RoI-free CondInst framework~\cite{eccv2020_condinst}.
DiscoBox~\cite{iccv2021discobox}  further introduces the intra-image and cross-image pairwise potential to build a self-ensembled  teacher network, where the pseudo-labels generated from the teacher are utilized by the student task network to reduce the uncertainties.
Despite of the promising performance, the pairwise affinity relationship is built upon either partial or all neighbouring pixel pairs with the oversimplified assumption that similar pixel or color pairs share the same label. This inevitably introduces much label noises, especially for complicated background or nearby similar objects.

Some recently developed methods~\cite{cvpr2021bbam, cvpr2021boxcaseg} focus on the generation of proxy mask as the pseudo-label supervision by using detachable networks. BBAM~\cite{cvpr2021bbam} and BoxCaseg~\cite{cvpr2021boxcaseg} employ multiple training stages or extra saliency data to achieve this goal, where the first stage is designed for pseudo mask generation. In addition to box annotation, Cheng \textit{et al.}~\cite{cvpr2022pointly} and Tang \textit{et al}.~\cite{tang2022active} proposed to employ extra points as supervision within the bounding box to obtain more accurate results.
Unlike the above methods, our proposed level-set based approach learns to evolve the object boundary curves implicitly in an end-to-end manner, which is able to iteratively segment the instances' boundaries by optimizing the energy function within the given bounding box region.

\subsection{Level-set based Image Segmentation}
As a classical variational approach, the level-set methods~\cite{levelset1995a, osher1988fronts} have been widely used in image segmentation, which can be categorized into two major groups: region-based approaches~\cite{tip2001_active_contour, ijcv2002multiphase, mumford1989optimal} and edge-based approaches~\cite{ijcv1997geodesic, tpami1995shape}. The key idea of level-set methods is to represent the implicit curves by an energy function in a higher dimension, which is iteratively minimized by using the gradient descent technique. 

Some works~\cite{eccv2020levelset,tip2019mumford, CVPR2017hudeep,cvpr2019deeplevelsetevolution,yuan2020deep} have been developed to embed the level-set evolution into the deep learning framework in an end-to-end manner so that strong high-level features can be utilized for curve evolution and promising segmentation results have been achieved. For example, 
Wang~\textit{et al.}~\cite{cvpr2019deeplevelsetevolution} predicted the evolution parameters to evolve the predicted contour by incorporating the user clicks on the extreme boundary points, where the edge-based level-set energy function~\cite{ijcv1997geodesic} is adopted. To facilitate the accurate instance segmentation, Levelset R-CNN~\cite{eccv2020levelset} performs the Chan-Vese level-set evolution with the deep features extracted by Mask R-CNN~\cite{iccv2017maskrcnn}, where the low-level features from the original image are not utilized in the optimization.
Yuan~\textit{et al.}~\cite{yuan2020deep} built a piecewise-constant function to parse each sub-region corresponding to an individual instance based on the Mumford–Shah model~\cite{mumford1989optimal}, which accomplishes instance segmentation by a fully convolutional neural network. These methods learn to evolve the level-set curves by using ground-truth mask in a fully supervised manner. In contrast, our proposed approach learns the level-set evolution by  using only the box annotations, free of the expensive pixel-wise mask annotations. 

Kim~\textit{et al.}~\cite{tip2019mumford} learned to perform level-set evolution in an unsupervised manner, which is the mostly related work to our approach. To achieve $N$-class semantic segmentation, they employed the global multi-phase Mumford–Shah function~\cite{mumford1989optimal} and evolved it only on the low-level features of input image. In contrast, our method is based on the Chan-Vese functional~\cite{tip2001_active_contour}, which is constrained within the local bounding box with the enriched information from both input image and high-level deep features. Moreover, the initialization of level-set is generated automatically in our approach for robust curve evolution.

\begin{figure*}[t]
	\centering
	\includegraphics[width=0.95\linewidth]{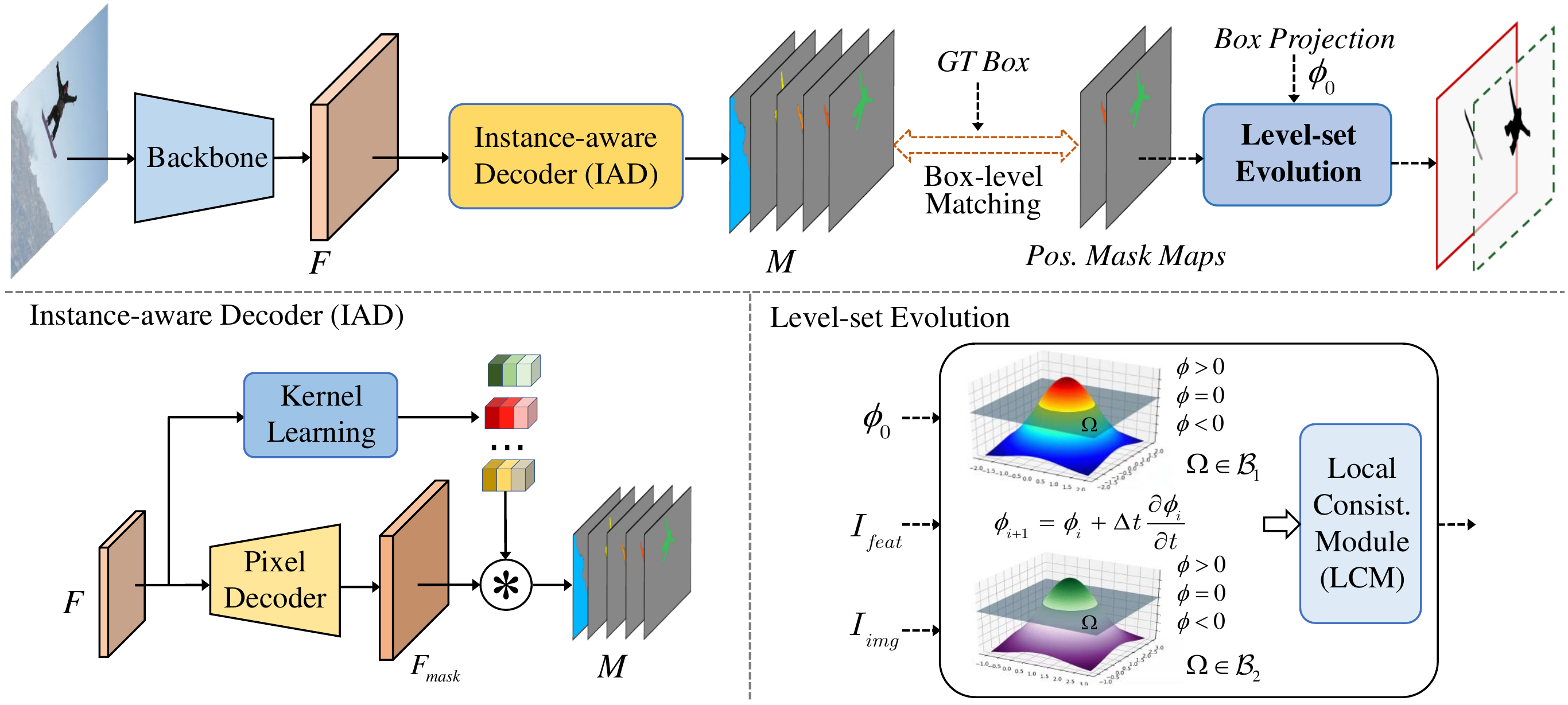}
	\caption{\textbf{Overall framework of Box2Mask.} 
		Our proposed Box2Mask is a single-stage method for box-supervised instance segmentation. It consists of a backbone, an instance-aware decoder (IAD), a box-level matching assignment step and a level-set evolution module. The backbone is adopted as the feature encoder to extract basic features.
		The IAD generates the instance-aware mask maps $M$ of full image size.   
		Only the positive mask maps of $M$ will perform level-set evolution within the corresponding bounding box region $\mathcal{B}$. 
		With iterative energy minimization, accurate instance segmentation can be obtained with box annotations only. The category branch is omitted here for the convenience of illustration.}
	\label{fig:overallnetwork}
\end{figure*}

\section{Proposed Method}

\subsection{Overview}
Fig.~\ref{fig:overallnetwork} provides the overview of our proposed Box2Mask method for box-supervised instance segmentation, which consists of a backbone, an instance-aware decoder (IAD), a box-level matching assignment step and a level-set evolution module.
Given an input image, a backbone is adopted as the feature encoder to extract the basic features.
Then the IAD learns to embed the characteristics of each instance by kernel learning to generate the instance-aware mask maps dynamically conditioned on the input instances. 
The box-level matching assignment is introduced to assign the high-quality mask map samples as the positives. Finally, we introduce a novel level-set evolution module to generate the accurate supervisions  with only bounding box annotations.
The designed level-set energy function enables the whole network to learn a series of level-set functions evolving to the instance boundaries implicitly.
We develop two types of frameworks, a CNN-based one and a transformer-based one, to ensure efficient level-set evolution in a single-stage manner. In each framework, high-quality mask prediction of an instance can be obtained via the iteratively optimization within its corresponding bounding box region.

In the following sections, we introduce the key components of our frameworks, including IAD, matching assignment and level-set evolution, in detail.

\subsection{Instance-aware Decoder}

The IAD learns to embed the unique characteristics (\textit{e.g.}, intensity, appearance, shape and location, \textit{etc}.) of each instance to generate the instance-aware mask map.  It mainly consists of a pixel-wise decoder and a kernel learning network, which decouples the instance-aware mask map output into the generation of unified mask features and the corresponding unique kernels.

\textbf{CNN-based IAD.}
For CNN-based IAD, we adopt the dynamic convolution method as in SOLOv2~\cite{wang2020solov2, PMAI2021solo}.
Based on the basic features $F$ extracted from the backbone, the  kernel learning network first employs several convolution layers to embed instance-wise characteristics and generate the instance-unique kernels $K_{i,j}$.
As the parallel branch, the pixel-wise decoder employs the Feature Pyramid Network (FPN)~\cite{cvpr2017_FPN} to extract multi-scale features, based on which the feature aggregation is performed at the same resolution to get the unified feature representation $F_{mask}$.
The learning kernel $K_{i,j}$ performs the convolution dynamically on the unified mask features $F_{mask}$ to generate the instance-aware mask map $M$, \textit{i.e.,} $M_{i,j} = K_{i,j} * F_{mask}$, where ${M_{i,j}}$ is the full-image mask map containing only one instance centered at location $(i,j)$. This CNN-based module produces the dynamic mask maps and distinguishes individual instances.

\begin{figure}[t]
	\centering
	\includegraphics[width=0.49\linewidth]{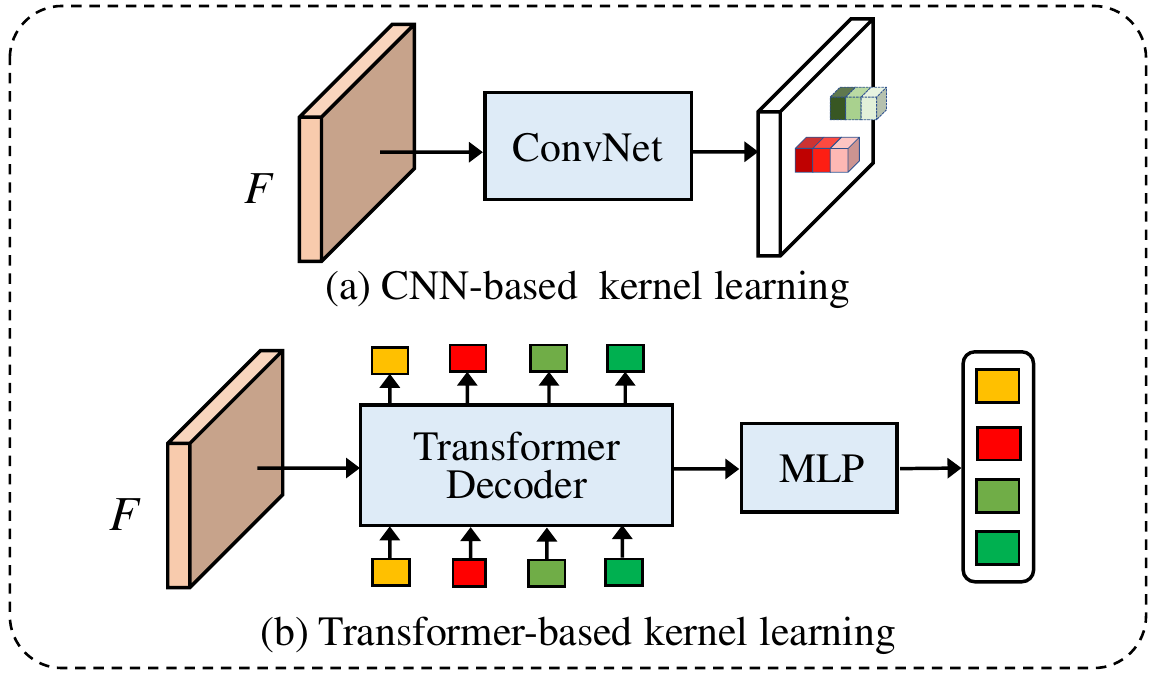}
	\caption{Kernel learning network in our Box2Mask framework. In the (a) CNN-based framework, the convolutional layers is employed in the kernel learning network. In the (b) transformer-based framework, the transformer decoder is adopted as the kernel learning network.}
	\label{fig:kernellearning}
\end{figure}

\textbf{Transformer-based IAD.}
Inspired by MaskFormer~\cite{nips2021maskformer}, 
the segmentation map of each instance can be represented as a $D$-dimensional feature vector, which can be computed by a transformer decoder~\cite{nips2017attention} with $N$ learnable instance-wise queries using the set prediction mechanism~\cite{eccv2020-detr}.
As shown in Fig.~\ref{fig:kernellearning}, 
to achieve dynamic IAD, we adopt a transformer decoder to compute $N$ instance-aware kernel vectors $K\in{\mathbb{R}^{N \times C}}$, which encode the object characteristics conditioned on input instances.
At the same time, a pixel decoder is used to gradually upsample the low-resolution features from the output of backbone to generate the high-resolution mask features $F_{mask}\in{\mathbb{R}^{C \times H \times W}} $. Specially, instead of the convolutional network, we adopt the multi-scale deformable attention transformer~\cite{zhu2020deformable} layers to extract stronger feature representation with long-range context.
The final instance-aware mask maps are decoded from the embeddings using object queries via a dot product operation with learned instance-aware kernels, \textit{i.e.,} 
${M^{N \times H \times W}} = {K^{N \times C}} \cdot F_{mask}^{C \times H \times W}$.  The total number of object queries in the transformer decoder is set to $N=100$ by default, and the object query embeddings are initialized as zero vectors with the learnable positional encoding.
With this transformer-based IAD, the set of instance-aware mask maps can be obtained in our transformer-based framework.

\subsection{Box-level Matching Assignment} 
Label assignment plays a critical role in training instance segmentation networks. It assigns the potential mask samples as the positives, and the reminder as negatives. With bounding box annotations only, we employ different matching assignment techniques for CNN-based and transformer-based frameworks to assign the category label and generate the positive instance-aware mask maps for the subsequent level-set evolution module (see Sec.~\ref{section3.3}).

For CNN-based framework, dense instance-aware map candidates will be generated at any location $(i,j)$ of an image. We employ the \textit{box-based center sampling} scheme to ensure that each potential instance-aware map contains only one target instance, whose center is at location $(i,j)$. The mask map candidate is considered as a positive sample if the corresponding location $(i,j)$ falls into the pre-defined center region of any ground truth bounding box; otherwise, it is treated as a negative sample. The area of center region is controlled by a scaling factor to ensure that on average there are 3 positive samples for each ground truth box ~\cite{wang2020solov2}.

For transformer-based framework, we adopt the Hungarian algorithm based bipartite matching assignment scheme~\cite{eccv2020-detr}, which matches between the instance-aware mask maps generated from IAD and the ground truth bounding boxes. 
The matching cost consists of both the category and spatial location differences between the class-wise mask predictions and ground truth bounding boxes.
For the matching cost $\mathcal{C}_{inst}$ on instance segmentation, we employ the coordinate projection $\mathcal{P}$ on $x$-axis and $y$-axis to calculate the spatial differences between the predicted instance-aware maps $m^p$  and box-based ground truth masks $m^b$ measured by 1-D dice coefficient~\cite{ic3dv2016_dice_loss}, \textit{i.e.,} $\mathcal{C}_{inst} =\mathcal{P}_{{dice}}({m^p_x}, m^b_x) +\mathcal{P}_{{dice}}({m^p_y}, m^b_y)$. 
Such a simple box-based projection cost function can reflect the spatial relationships accurately.
Besides, the commonly used cross entropy loss is adopted as the category cost ${\mathcal{C}_{cate}}$  to measure the similarity of predicted category and ground-truth class label. Therefore, the total matching cost $\mathcal{C}$ can be formulated as follows: 
\begin{equation}\label{mathing_cost_eq}
	\mathcal{C} = \beta  {}_1 \mathcal{C}_{inst} + \beta {}_2 \mathcal{C}_{cate}.
\end{equation}
The instance-aware mask maps with no match will yield a ``no object" ($\emptyset$) target class label. With the above matching assignment scheme, we can guarantee the unique one-to-one matching during the transformer model training, and predict a single mask map for each box-based ground truth.

\subsection{Level-set Evolution} 
\label{section3.3}

We firstly introduce the formulation of level-set model for segmentation, then provide the details of the level-set evolution module in our Box2Mask method.

\subsubsection{Level-set Model}
As a classical approach, the level-set method~\cite{levelset1995a, osher1988fronts} has been widely used for image segmentation. It formulates the segmentation as a consecutive energy minimization problem~\cite{mumford1989optimal, tip2001_active_contour, ijcv2002multiphase}.
In the Mumford-Shah level-set model~\cite{mumford1989optimal}, the segmentation of a given image $I$ is obtained by finding a parametric contour $C$, which partitions the image plane $\Omega \subset \mathbb{R}^2$ into $N$ disjoint regions ${\Omega _1}, \cdots ,{\Omega _N}$. The Mumford-Shah energy functional $\mathcal{F}^{MS}(u ,C)$ can be written as below:
\begin{normalsize}
	\begin{equation} \label{eq2}
		\begin{aligned}
			{\mathcal{F}^{MS}}({u_1}, \! \cdots, \! {u_N}, \! {\Omega _1}, \! \cdots , \! {\Omega _N}) 
			 = \sum\limits_{i = 1}^N {(\! \int\limits_{{\Omega _i}} {{{(\! I - \! {u_i})}^2}} dxdy + \! \mu \! \int\limits_{{\Omega _i}} {{{\left| {\nabla {u_i}} \right|}^2}dxdy \! + \! \gamma \! \left| {{C_i}} \right|})}, 
		\end{aligned}
	\end{equation}
\end{normalsize}

\noindent where $u_i$ is a piecewise smooth function approximating the input image $I$, ensuring the smoothness inside each region ${\Omega _i}$. $\mu$ and $\gamma$ are the weighting parameters.

Chan and Vese~\cite{tip2001_active_contour} simplified the Mumford-Shah functional as a variational level-set, which has been well explored~\cite{wang2010efficient,liu2012local,mavska2013segmentation,xu2011image}. Specifically, it can be written as follows:
\begin{normalsize}
	\begin{equation} \label{eq3}
		\begin{aligned} 
			{\mathcal{F}^{CV}}(\phi ,{c_1},{c_2}) &= \int\limits_\Omega  {{{\left| {I(x,y) - {c_1}} \right|}^2}H(\phi (x,y))} dxdy \\ & + \int\limits_\Omega  {{{\left| {I(x,y) - {c_2}} \right|}^2}(1 - H(\phi (x,y)))} dxdy  +  \gamma \int\limits_\Omega  {\left| {\nabla H(\phi (x,y))} \right|dxdy} 
		\end{aligned}
	\end{equation}
\end{normalsize}

\noindent
where $H$ is the Heaviside function, and $\phi(x,y)$ is the level-set function, whose zero crossing contour $C = \{ (x,y):\phi (x,y) = 0\}$ divides the image space $\Omega$ into two disjoint regions, 
inside contour $C$: ${\Omega _1} = \{ (x,y):\phi (x,y) > 0\}$ and outside contour $C$: ${\Omega _2} = \{ (x,y):\phi (x,y) < 0\}$. 
In Eq.~(\ref{eq3}), the first two terms intend to fit the data, and the last term regularizes the zero level contour with a non-negative parameter $\gamma$. $c_1$ and $c_2$ are the mean values of input $I(x,y)$ inside $C$ and outside $C$, respectively. The image segmentation is achieved by finding a level-set function $\phi(x,y)= 0$ with ${c_1}$ and ${c_2}$ that minimizes the energy $\mathcal{F}^{CV}$.

\subsubsection{Level-set Evolution within Bounding Box}
Our proposed method exploits the level-set evolution with Chan-Vese energy-based model~\cite{tip2001_active_contour} to achieve high-quality instance segmentation.
Firstly, we treat each positive instance-aware mask map as the level-set function $\phi$ of its assigned object with bounding box annotation. Both low-level input image feature $I_{img}$ and high-level deep features $I_{feat}$ are utilized as the input data terms to evolve the level-set, while a box projection function encourages the network to automatically estimate an initial level-set $\phi_0$ at each step. 
The level-set for each instance is iteratively optimized within its corresponding bounding box region so that high-quality mask predictions can be obtained.

\textbf{Level-set Evolution.}  Given an input image $I(x,y)$, we aim to predict the object boundary curve by evolving a level-set implicitly within the region of annotated bounding box $\mathcal{B}$.
The instance-aware mask map $M \in {\mathbb{R}^{H \times W \times N}}$  contains $N$ potential instance maps of size $H \times W$.   
After the box-level matching  assignment, the positive and negative mask map samples are split. 
We treat each positive mask map within box $\mathcal{B}$ as the level-set $\phi(x,y)$, and its corresponding pixel space of input data $I(x,y)$ is referred as $\Omega$, \textit{i.e.}, $\Omega  \in \mathcal{B}$. $C$ is the segmentation boundary with zero level $C = \{ (x,y):\phi (x,y) = 0\}$, which partitions the box region into two disjoint regions, i.e., foreground object and background. 

To obtain the accurate boundary for each instance, we learn a series of level sets $\phi(x,y)$ by minimizing the following energy function:
\begin{equation} \label{eq4}
	\begin{aligned}
		\mathcal{F}(\phi ,I, {c_1},{c_2},\mathcal{B}) &=  \int\limits_{\Omega  \in \mathcal{B}} {{{\left| {{I^*}(x,y) - c_1} \right|}^2}\sigma (\phi (x,y))} dxdy \\ & +  \int\limits_{\Omega  \in \mathcal{B}} {{{\left| {{I^*}(x,y) - c_2} \right|}^2}(1 - \sigma (\phi (x,y)))} dxdy  +  \gamma \int\limits_{\Omega  \in \mathcal{B}} {\left| {\nabla \sigma (\phi (x,y))} \right|dxdy} 
	\end{aligned}
\end{equation}
where $I^*(x,y)$ denotes the normalized input image $I(x,y)$.
$\gamma$ is a non-negative weight, and
$\sigma$ denotes the \texttt{sigmoid} function that is treated as the characteristic function for level-set $\phi(x,y)$.  Different from the traditional Heaviside function~\cite{tip2001_active_contour}, the \texttt{sigmoid} function is much smoother, which can better express the characteristics of the predicted instance and improve the convergence of level-set evolution during the training process. The first two items in Eq.~(\ref{eq4}) force the predicted $\phi(x,y)$
to be uniform both inside region $\Omega$ and outside region
$\bar{\Omega}$. $c_1$ and $c_2$ are the mean values of $\Omega$ and $\bar{\Omega}$, which are defined as below:
\begin{equation} \label{eq5}
	\begin{aligned}
		c_1(\phi ) = \frac{{\int\limits_{\Omega  \in \mathcal{B}} {{I^*}(x,y)\sigma (\phi (x,y))} dxdy}}{{\int\limits_{\Omega  \in \mathcal{B}} {\sigma (\phi (x,y))} dxdy}},  \ \ \
	    c_2(\phi ) = \frac{{\int\limits_{\Omega  \in \mathcal{B}} {{I^*}(x,y)(1 - \sigma (\phi (x,y)))} dxdy}}{{\int\limits_{\Omega  \in \mathcal{B}} {(1 - \sigma (\phi (x,y)))} dxdy}}.
	\end{aligned}
\end{equation}

The energy function $\mathcal{F}$ can be optimized with  gradient back-propagation during training. With the time step $t \geqslant 0$, the derivative of energy function $\mathcal{F}$ upon $\phi$ can be written as follows:
\begin{equation} \label{eq7}
	\begin{aligned}
		\frac{{\partial \phi }}{{\partial t}} =  - \frac{{\partial \mathcal{F}}}{{\partial \phi }}  =  - \nabla \sigma (\phi )[ {({I^*}(x,y) - c_1)^2} - {({I^*}(x,y) - c_2)^2} + \gamma \texttt{div}\left (\frac{{\nabla \phi }}{{\left| {\nabla \phi } \right|}}\right)],
	\end{aligned}
\end{equation}
where $\nabla$ and $\texttt{div}$ are the spatial derivative and  divergence operators, respectively. Therefore, the update of $\phi$ can be computed by
\begin{equation} \label{eq9}
	\begin{aligned}
		{\phi _n} = {\phi _{n - 1}} + \Delta t\frac{{\partial \phi_{n-1} }}{{\partial t}}.
	\end{aligned}
\end{equation}

The minimization of the above terms can be viewed as an implicit curve evolution along the descent of energy function. The optimal boundary $C$ of the instance is obtained by iteratively fitting $\phi _i$ as follows:
\begin{equation} \label{eq10}
	\begin{aligned}
		\mathop {\inf }\limits_{\Omega \in \mathcal{B}} \{ \mathcal{F}(\phi)\}  \approx 0 \approx \mathcal{F}({\phi_{i}}).
	\end{aligned}
\end{equation}

\textbf{Input Data Terms.} The energy function in Eq.~(\ref{eq4}) encourages the curve evolution based on the uniformity of regions inside and outside the object. The input image $I_u$ represents the basic low-level features, including shape, colour, image intensities, etc. However, such low-level features usually vary with different illuminations, different materials and motion blur, making the level-set evolution less robust.

In addition to the normalized input image, we take into account the high-level deep features $I_f$, which embed the image semantic information, to obtain more robust results. To this end, we make use of the unified mask feature $F_{mask}$ from all scales in the IAD module, and feed it into a convolution layer to extract the high-level features $I_f$. 
The features $I_f$ are further transformed by the tree filter~\cite{nips2019learnable, liang2022tree}, which employs the minimal spanning tree (MST) to model long-range dependencies and preserve the object's structure. 
The overall energy function for level-set evolution can be formulated as follows:
\begin{equation} \label{eq8}
	\begin{aligned}
		\mathcal{F(\phi)}\! =\! \lambda_1\! * \! {\mathcal{F}}(\phi,\! I_u,\! c_{{u_1}},\! c_{{u_2}},\! \mathcal{B}) \! +\! \lambda_2 *  {\mathcal{F}}(\phi,\! I_f, \! c_{{f_1}}, \! c_{{f_2}},\! \mathcal{B}),
	\end{aligned}
\end{equation}
where $\lambda_1$ and $\lambda_2$ are weights to balance the two kinds of features. $c_{{u_1}}$, $c_{{u_2}}$ and $c_{{f_1}}$, $c_{{f_2}}$ are the mean values for input data terms $I_u$ and $I_f$, respectively.

\textbf{Level-set Initialization.} Conventional level-set methods are sensitive to the initialization, which is often manually defined. In this work, we employ a box projection function~\cite{cvpr2021_boxinst} to enable the model to automatically generate a rough estimation of the initial level-set $\phi_0$ at each step.

In particular, we utilize the same coordinate projection function $\mathcal{P}_{{dice}}$ as in box-level matching assignment, and calculate the projection difference between the predicted mask map and the ground-truth box on $x$-axis and $y$-axis. 
This scheme limits the predicted initialization boundary within the bounding box, providing a good initial state for curve evolution. 
Let $m^b \in {\{ 0,1\} ^{H \times W}}{\rm{ }}$ denote the binary region by assigning 1 to the locations in the ground-truth box, and 0 otherwise. The mask score predictions ${m^p} \in {(0,1)^{H \times W}}$ for each instance can be regarded as the probabilities of foreground. The box projection function ${\mathcal{F}(\phi_0)_{box}}$ is defined as below:
\begin{equation} \label{eqbox}
	\begin{aligned}
		{\mathcal{F}(\phi_0)_{box}} = \mathcal{P}_{dice}({m^p_x},{m^b_x}) + \mathcal{P}_{dice}({m^p_y},{m^b_y}),
	\end{aligned}
\end{equation}
where ${m^p_x}$, $m^b_x$ and ${m^p_y}$, $m^b_y$ denote the $x$-axis projection and $y$-axis projection for mask prediction $m^p$ and binary ground-truth region $m^b$, respectively. 

\subsubsection{Local Consistency Module}
The region-based level-set for segmentation typically relies on the intensity homogeneity of the input data~\cite{CVPR2007implicit, tip2011level, tip2019mumford}. 
To alleviate the intensity inhomogeneity brought by region-based level-set,
we present an affinity kernel function to ensure the local consistency of level-set $\phi$, which further mines the pixel similarities 
and spatial relationships within the neighborhood.

First, we define an affinity kernel function $\mathcal{A}$ on the pixel intensities $p_{i,j}$ with 8-way local neighbors, which can be expressed as follows:
\begin{small}
	\begin{equation}
		\mathcal{A}_{_{i,j}}^p = \mathop {\mathcal{A} \left( {{p_{i,j}},{p_{l,k}}} \right)}\limits_{(l,k) \in \mathcal{N}_8(i,j)}  =  - {\left( {\frac{{\left\| {{p_{i,j}} - {p_{l,k}}} \right\|}}{{\sigma _{_{i,j}}^p}}} \right)^2},
	\end{equation}
\end{small}

\noindent 
where ${\sigma _{i,j}}$ denotes the standard deviation of the pixel intensity, computed locally for the affinity kernel. Then, the \texttt{softmax} function is employed to normalize the affinity distance $\mathcal{A}_{_{i,j}}^p$ on the pixel intensity:
\begin{small}
	\begin{equation}
		{\mathcal{A} {_{i,j}^p}^*} = \frac{{\exp (\mathcal{A} _{_{i,j,l,k}}^p)}}{{\sum\limits_{(l,k) \in \mathcal{N}_8(i,j)} {\exp (\mathcal{A} _{_{i,j,l,k}}^p)} }}.
	\end{equation}
\end{small}

\noindent Besides, we apply this affinity kernel function on the spatial location to encode the spatial relationship ${\mathcal{A}{_{i,j}^s}^*}$ within the neighborhood~\cite{ru2022learning}. Therefore, the whole affinity kernel function is defined as follows:
\begin{equation} \label{affinity}
	{\mathcal{A} _{i,j}}^* = {\mathcal{A} {_{i,j}^p}^*} + \eta \mathcal{A} {_{i,j}^s}^*,
\end{equation}
where $\eta$ is the balance weight.

To fully exploit the pixel and spatial affinity and keep the local consistency of level-set predictions, we perform this operation for $k$ times on the predicted level-set $\phi_n$ in each training step $t$:
\begin{equation}
	\begin{aligned}
		\phi _{i,j}^k = \sum\limits_{\Omega \in \mathcal{B}} {{\mathcal{A} _{i,j}}}^*  * \phi _{i,j}^{k - 1}.
	\end{aligned}
\end{equation}

\noindent 
In our implementation, we set $k=10 $ by default. By exploiting the pixel similarity and position relationship, robust level-sets with local affinity consistency can be obtained. We employ the simple $L_1$ distance between the predicted level-set $\phi _n$ and the refined $\phi _{n}^k$ to make  $\phi _n$ more stable and suitable to fit the target boundary.

\subsection{Training Loss and Inference}
We employ the level-set energy as the training objective for network optimization in an end-to-end fashion. Once the network is trained, the inference stage is efficient, which could directly output the mask prediction without the iterative level-set evolution process.

\textbf{Loss Function.} 
The loss function to train our Box2Mask network consists of two items, the category classification loss $L_{cate}$ and instance segmentation loss $L_{inst}$: 
\begin{equation} \label{loss}
	\begin{aligned}
		L = w{L_{cate}} + {L_{inst}},
	\end{aligned}
\end{equation}
where $w$ is the balance parameter. Here $L_{cate}$ can be instantiated as the commonly used cross-entropy loss or Focal loss~\cite{lin2017focal}. 
For $L_{inst}$, we employ the differentiable level-set energy as the objective: 
\begin{equation} \label{levelsetloss}
	\begin{aligned}
		{L_{inst}} = \frac{1}{{{N_{pos}}}}\sum\limits_k {{\mathbbm{1}_{\{\text{cond}\}}}\{ }    \alpha \mathcal{F}{({\phi _0})_{box}} +  \mathcal{F}(\phi )\}.
	\end{aligned}
\end{equation}
where $N_{pos}$ indicates the number of positive samples and $\alpha$ is the balance weight. $\mathbbm{1}_{\{\text{cond}\}}$ represents the indicator function with the corresponding conditions in CNN-based and transformer-based frameworks, which ensures that only the positive instance mask samples will be involved in the level-set evolution.
In the CNN-based framework,  $p_{i,j}^*$ is the category probability of target located at $(i,j)$, and hence the conditional function is $p_{i,j}^* > 0 $ to filter the low-quality candidates.
In the transformer-based framework, the target class $c_i$ should not be empty. Therefore, the conditional function is  $c_j^{gt} \ne \emptyset$. With the above loss function, our Box2Mask model can be trained in an end-to-end fashion.

\textbf{Inference.}
It is worth noting that \textit{the level-set evolution is only employed during training} to generate implicit supervisions for the network optimization. 
The inference process is direct and efficient without the need of level-set evolution.
Given the input image, the instance-wise mask predictions are directly generated. For the model trained under the CNN-based framework, the efficient matrix non-maximum suppression (NMS)~\cite{PMAI2021solo} is needed for post-processing. For the model trained under the transformer-based framework, the inference process is NMS-free, which directly outputs the instance masks.

\section{Experiments}

\subsection{Datasets}
To evaluate our proposed Box2Mask approach, we conduct extensive experiments on five challenging datasets, including Pascal VOC~\cite{pascalvoc2010} and COCO~\cite{lin2014microsoft}, remote sensing dataset iSAID~\cite{cvpr2019isaid}, medical image dataset LiTS~\cite{bilic2019lits} and scene text image dataset ICDAR2019 ReCTS~\cite{zhang2019icdar}.

\textbf{Pascal VOC}~\cite{pascalvoc2010}. Pascal VOC consists of 20 common categories. As in~\cite{nips2019-bbtp, cvpr2021_boxinst, iccv2021discobox}, the augmented Pascal VOC 2012~\cite{iccv2011_SBDdataset} dataset is used here, which contains 10,582 images for training and 1,449 images for evaluation. 

\textbf{COCO}~\cite{lin2014microsoft}. COCO has 80 general object classes. As a common setting, our models are trained on \texttt{train2017} (115K images), and evaluated on \texttt{val2017} (5K images) and \texttt{test-dev} split (20K images).

\textbf{iSAID}~\cite{cvpr2019isaid}. iSAID is a large-scale high-resolution remote sensing dataset for aerial instance segmentation, containing many small objects with complex backgrounds. It has 15 categories and comprises 1,411 images for training and 458 images for performance evaluation with 655,451 instance annotations in total. 

\textbf{LiTS}~\cite{bilic2019lits}. The Liver Tumor Segmentation Challenge (LiTS) dataset$\footnote{https://competitions.codalab.org/competitions/17094}$ consists of 130 volume CT scans for training and 70 volume CT scans for testing. We randomly partition all the scans having mask labels into the training and validation dataset with a ratio of 4:1.

\textbf{ICDAR2019 ReCTS}~\cite{zhang2019icdar}. This scene text image dataset contains 20K images for training and 5K images for testing. The training images are annotated with character-level boxes and text-lines. Due to the lack of annotations in testing set, we randomly divide the training set into the training and validation subsets with a ratio of 4:1.

\subsection{Implementation Details}
Our models are trained with the AdamW optimizer~\cite{AdamW2017decoupled} on 8 NVIDIA V100 GPUs. 
We use the \texttt{mmdetection} toolbox~\cite{chen2019mmdetection} and follow the commonly used training settings on each dataset. ResNet~\cite{he2016deep} and Swin-Transformer~\cite{liu2021swin} are employed as the backbones, which are pre-trained on ImageNet-1K~\cite{IJCV2015imagenet}.
For the models trained under the CNN-based framework, denoted by Box2Mask-C, the initial learning rate is $10^{-4}$ and the weight decay is $0.1$ with 16 images per mini-batch. The training schedules of ``1$\times$" and ``3$\times$" are the same as \texttt{mmdetection} with 12 epochs and 36 epochs, respectively. In the loss function of Eq.~\ref{levelsetloss},  we set $\alpha=3.0$.
For the models trained under the transformer-based framework, denoted by Box2Mask-T, the initial learning rate is set to $5\times10^{-5}$ and the weight decay is $0.05$ with 8 images per mini-batch. As in ~\cite{cvpr2022mask2former}, we train the models for $50$ epochs, and the large-scale jittering augmentation scheme~\cite{cvpr2021simple} is employed with a random scale sampled within range [0.1, 2.0], followed by a fixed size crop to 1024$\times$1024.  We set $\alpha=5.0$ in Eq.~\ref{levelsetloss} for our Box2Mask-T models.
The non-negative weight $\gamma$ in Eq.~\ref{eq4} is set to $10^{-4}$. To keep the level-set energy value at the same level for input data terms (including original image and high-level deep features), we set ${\lambda _1} = 0.05 $ and ${\lambda _2} = 5.0 $ in Eq.~\ref{eq8}, respectively.

On COCO and Pascal VOC, the scale jitter is used,  where the shorter image side is randomly sampled from 640 to 800 pixels for a fair comparison. On the iSAID dataset, all models are trained with the input size of 800$\times$800 unless otherwise specified. On the LiTS dataset, all models are trained with 640$\times$640 input size.
On ICDAR 2019 ReCTS,  all the training settings are the same as that of COCO with character-level boxes annotations.
COCO-style mask AP ($\%$) is adopted for performance evaluation. Like~\cite{cvpr2021bbam, iccv2021discobox}, we report the average precision (AP) at four IoU thresholds (including 0.25, 0.50, 0.70 and 0.75) and the average best overlap (ABO) for the comparison on Pascal VOC dataset.

\subsection{Main Results}

We compare our proposed Box2Mask method against the state-of-the-art box-supervised instance segmentation approaches. The results of representative fully mask-supervised methods are also reported for reference.

\begin{table*}[t]
	\renewcommand\arraystretch{0.94}
	\centering
	\caption{Performance comparison on Pascal VOC \texttt{val}. ``$*$" denotes the results of GrabCut~\cite{TOG2004grabcut} reported from  BoxInst~\cite{cvpr2021_boxinst}. 
	All the results are obtained using only \textit{box supervision}. Backbone pre-trained on ImageNet-22K is marked with $^\triangleleft$ (the same in other tables).} 
	\scalebox{0.9}{
	\setlength{\tabcolsep}{1.7mm}{
		\begin{tabular}{lllcccccc}
			\toprule
			Method  & Pub.& Backbone  &AP & AP$_{25}$  &  AP$_{50}$ & AP$_{70}$& AP$_{75}$ & ABO~ \\
			\midrule
			GrabCut$^*$~\cite{TOG2004grabcut} &TOG'04 & ResNet-101 &19.0 &- & 38.8 & - & 17.0 & -~ \\
			SDI~\cite{cvpr2017SDI} &CVPR'17 & VGG-16 &- & -  &44.8 & - & 16.3  & 49.1~ \\
			Liao \textit{et al.}~\cite{liao2019weakly} & IEEE ASSP'19 & ResNet-101  & - & - & 51.3 & - & 22.4  & 51.9~ \\
			Sun \textit{et al.}~\cite{sun2020weakly}  & IEEE Access'20 & ResNet-50  & - &- & 56.9 & - & 21.4 & 56.9~ \\ 
			
			BBTP~\cite{nips2019-bbtp}  &NeurIPS'19 & ResNet-101& 23.1 & -& 54.1 & -& 17.1 & -~ \\
			BBTP w/ CRF~\cite{nips2019-bbtp}  & NeurIPS'19 & ResNet-101 & 27.5 & - & 59.1 & - & 21.9 &-~ \\
			Arun \textit{et al}.~\cite{arun2020weakly}  &ECCV'20 & ResNet-101 &  - & 73.1 &57.7 & 33.5 & 31.2 & -~ \\
			BBAM~\cite{cvpr2021bbam}  &CVPR'21 & ResNet-101 &- &76.8 & 63.7  & 39.5 & 31.8 & 63.0~ \\
			
			A$^2$GNN w/ CRF~\cite{TPAMI2021affinity} & TPAMI'21  & ResNet-101 & - & - & 59.1 & 35.5 & 27.4 &-~\\ 
			
			Zhang~\textit{et al.}~\cite{zhang2022weakly} & PR'22 & ResNet-50 & 34.3 & - & 58.6 & - & 34.6 & -~ \\
			Zhang~\textit{et al.}~\cite{zhang2022weakly} & PR'22 & ResNet-101 & 36.4 & - & 60.9 & - & 37.7 & -~ \\

			BoxInst~\cite{cvpr2021_boxinst}  &CVPR'21 & ResNet-50 & 34.3 &- & 59.1 & -  & 34.2  & -~\\ 
			BoxInst~\cite{cvpr2021_boxinst}  &CVPR'21 & ResNet-101 & 36.5& - & 61.4 & - & 37.0 & -~ \\ 
			
			DiscoBox~\cite{iccv2021discobox} &ICCV'21  & ResNet-50-DCN &-& 75.2 &63.6 & 41.6& 34.1 &-~\\ 
			DiscoBox~\cite{iccv2021discobox}  &ICCV'21 & ResNet-50 & - & 71.4& 59.8& 41.7& 35.5 &-~\\ 
			DiscoBox~\cite{iccv2021discobox}  &ICCV'21 & ResNet-101 & - & 72.8 &62.2 & 45.5 & 37.5 &-~\\ 
			
			\hline
			\rowcolor{gray!8}
			Box2Mask-C   & - & ResNet-50 & 38.0 & 76.9& 65.9& 46.1 & 38.2  & 70.9~ \\
			\rowcolor{gray!8}
			Box2Mask-C & - & ResNet-101 & 39.6 & 77.3 & 66.6 & 47.9 & 40.9 & 71.9~ \\
			
			\rowcolor{gray!8}
			Box2Mask-T  & - & ResNet-50 & 41.4 & 81.5 & 68.9  & 49.3 & 42.1 & 73.9  \\
			\rowcolor{gray!8}
			Box2Mask-T  & - & ResNet-101 & 43.2 & 83.2 & 70.8 & 50.8 & 44.4 & 75.2~\\
			\rowcolor{gray!8}
			Box2Mask-T  & - & Swin-B$^\triangleleft$ & \textbf{48.9} & \textbf{88.3} & \textbf{77.2} & \textbf{57.8} & \textbf{51.1} & \textbf{77.6}~\\
			\bottomrule
	\end{tabular}}}
	\label{tab:vocsota}
\end{table*}

\begin{table*}[t]
	\renewcommand\arraystretch{0.92}
	\centering
	\caption{Performance comparison on COCO \texttt{test}-\texttt{dev}. ``$\dag$" denotes the result on the COCO \texttt{val2017} split. LIID$^\ddag$~\cite{tpami2020leveraging} needs  proposals generated by MCG~\cite{tpmai2017mcg} with only image-level annotations.   ``$*$" indicates that the BoxCaseg~\cite{cvpr2021boxcaseg} is trained with box and extra salient object supervisions. Both mask-supervised and box-supervised methods are compared here.} 
	\scalebox{0.9}{
	\setlength{\tabcolsep}{2.5mm}{
		\begin{tabular}{lllcccccc}
			\toprule
			Method & Pub.  & Backbone &AP & AP$_{50}$ & AP$_{75}$&AP$_{S}$ & AP$_{M}$ & AP$_{L}$~  \\
			\midrule
			\multicolumn{9}{c}{\small \em mask-supervised} \\
			\midrule
			Mask R-CNN~\cite{iccv2017maskrcnn}& CVPR'17 & ResNet-101 &35.7  &58.0   &37.8   &15.5   &38.1   & 52.4~ \\
			YOLACT-700~\cite{iccv2019yolact} & ICCV'19 &ResNet-101 &  31.2 &50.6& 32.8& 12.1& 33.3 & 47.1~   \\
			PolarMask~\cite{cvpr_2020polarmask} & CVPR'20 & ResNet-101  & 32.1&53.7 &33.1& 14.7 &33.8 & 45.3~ \\
			SOLO~\cite{PMAI2021solo} & TPAMI'21 & ResNet-101 & 37.8 &59.5& 40.4 &16.4& 40.6& 54.2~  \\
			SOLOv2~\cite{PMAI2021solo} &TPAMI'21 & ResNet-101 & 39.7 &60.7 &42.9& 17.3& 42.9 &57.4~\\
			CondInst~\cite{tian2022instance} &TPAMI'22 & ResNet-101  &  39.1 &60.9 & 42.0 & 21.5 & 41.7 & 50.9~ \\
			K-Net~\cite{nips2021knet} &NeurIPS'21 & ResNet-101 & 40.1 & 62.8 & 43.1 & 18.7 & 42.7 & 58.8~ \\
			QueryInst~\cite{ICCV2021_queryinst} & ICCV'21 & ResNet-101  & 41.7  & 64.4 & 45.3 & 24.2 & 43.9 & 53.9~ \\
			Mask2Former$^\dag$~\cite{cvpr2022mask2former} &  CVPR'22 & ResNet-101 & 44.2 & - & - & 23.8 & 47.7 & 66.7~\\
			\midrule
			\multicolumn{9}{c}{\small \em box-supervised} \\
			\midrule
			BBTP$^\dag$~\cite{nips2019-bbtp} & NeurIPS'19 & ResNet-101 & 21.1 &45.5 &17.2& 11.2 &22.0& 29.8~ \\
			
			Zhang~\textit{et al.}$^\dag$~\cite{zhang2022weakly} & PR'22 & ResNet-50 & 28.9 & 49.3 & 29.4 & - & - & - ~\\
			
			DiscoBox$^\dag$~\cite{iccv2021discobox} &ICCV'21 & ResNet-50 & 31.4 & 52.6 & 32.2 & 11.5 & 33.8 & 50.1~ \\
			
			\rowcolor{gray!8}
			Box2Mask-C$^\dag$ & - & ResNet-50 & 32.2 & 54.4 & 32.8 & 14.7& 35.9 & 47.5~ \\
			\rowcolor{gray!8}
			Box2Mask-T$^\dag$ & - & ResNet-50 & 36.1 & 60.8 & 36.4 & 16.2 & 38.5 & 56.6~  \\
			\midrule
			DiscoBox~\cite{iccv2021discobox} &ICCV'21 & ResNet-50 & 32.0 & 53.6 & 32.6 & 11.7 & 33.7 & 48.4~ \\
			
			BoxInst~\cite{cvpr2021_boxinst} & CVPR'21 & ResNet-50 & 32.1 & 55.1 & 32.4 & 15.6 & 34.3& 43.5~ \\
			
			\rowcolor{gray!8}
			Box2Mask-C & - & ResNet-50 & 32.6 & 55.4& 33.4 & 14.7& 35.8 & 45.9~ \\
			
			\rowcolor{gray!8}
			Box2Mask-T  & -&ResNet-50 & 36.7 & 61.9 & 37.2 & 18.2 & 39.6 & 53.2~ \\
			\midrule
			LIID$^\ddag$~\cite{tpami2020leveraging} & TPAMI'20 & ResNet-101 & 16.0 & 27.1 & 16.5 & 3.5 & 15.9 & 27.7~\\
			A$^2$GNN w/ CRF~\cite{TPAMI2021affinity} & TPAMI'21 & ResNet-101 & 20.9 & 43.9 & 17.8 & 8.3 & 20.1 & 31.8~\\
			
			BBAM~\cite{cvpr2021bbam} & CVPR'21 &ResNet-101 & 25.7 & 50.0& 23.3 & - & -& -~ \\ 
			
			BoxCaseg*~\cite{cvpr2021boxcaseg} &CVPR'21 & ResNet-101  &30.9 & 54.3 & 30.8 & 12.1 & 32.8 & 46.3~ \\
			
			BoxInst~\cite{cvpr2021_boxinst} & CVPR'21 & ResNet-101 & 33.2 & 56.5& 33.6& 16.2& 35.3& 45.1~ \\
			
			\rowcolor{gray!8}
			Box2Mask-C & - & ResNet-101 & 34.2 & 57.8 & 35.2 & 16.0 & 37.7 & 48.3~ \\
			\rowcolor{gray!8} 
			Box2Mask-T  & - & ResNet-101 & 38.3 & 65.1 & 38.8 & 19.3 & 41.7 & 55.2~  \\

			\rowcolor{gray!8}
			Box2Mask-T & -& Swin-B$^\triangleleft$ & 41.5 & 69.2 & 42.1 & 20.8 & 45.1 & 61.7~  \\
			\rowcolor{gray!8}
			Box2Mask-T & - & Swin-L$^\triangleleft$ & \textbf{42.4} & \textbf{70.2} & \textbf{43.3} & \textbf{22.1} & \textbf{45.9} & \textbf{62.9}~ \\
			\bottomrule
	\end{tabular}}}
	\label{tab:coco_results}
\end{table*}

\subsubsection{Instance Segmentation on General Scenes}

\noindent \textbf{Results on Pascal VOC.}  Table~\ref{tab:vocsota} reports the comparison results on Pascal VOC \texttt{val}. Our proposed Box2Mask with CNN-based framework (Box2Mask-C) outperforms BoxInst~\cite{cvpr2021_boxinst} by $3.7\%$  and $3.1\%$ mask AP with ResNet-50 and ResNet-101 backbones, respectively. 
Furthermore, Box2Mask with transformer-based framework (Box2Mask-T) obtains $43.2\%$ AP, significantly surpassing BoxInst~\cite{cvpr2021_boxinst} by \textbf{absolute 6.7\% mask AP} (\textit{e.g.}, from $36.5\%$ to $43.2\%$) with the same ResNet-101 backbone. 
For the AP$_{25}$ and AP$_{50}$ measures, Box2Mask-C achieves $77.3\%$ and $66.6\%$, which significantly outperforms the recent method DiscoBox~\cite{iccv2021discobox} by $4.5\%$ and $4.4\%$, respectively. 
For high IoU threshold-based AP metrics, Box2Mask-C achieves $40.9\%$ AP$_{75}$ with ResNet-101, which surpasses BoxInst~\cite{cvpr2021_boxinst} and DiscoBox~\cite{iccv2021discobox} by $3.9\%$ and $3.4\%$, respectively. 
The high IoU threshold-based AP metrics can more faithfully reflect the segmentation
performance with accurate boundary, which is in line with the practical applications.

Compared with Box2Mask-C, Box2Mask-T  achieves much higher performance. With ResNet-101 backbone, Box2Mask-T obtains $83.2\%$ AP$_{25}$, $70.8\%$ AP$_{50}$, $50.8\%$ AP$_{70}$,  $44.4\%$ AP$_{75}$ and $75.2\%$ ABO, outperforming Box2Mask-C by +$5.9\%$, +$4.2\%$,  +$2.9\%$, +$3.5\%$ and +$3.3\%$ accordingly.
Furthermore, with the Swin-Transformer base (Swin-B) model as the backbone, \textbf{Box2Mask-T obtains 48.9\% mask AP} on Pascal VOC. 
The outstanding results clearly demonstrate the great potentials of our box-supervised instance segmentation method. 

\textbf{Results on COCO.} 
Table~\ref{tab:coco_results} compares Box2Mask against state-of-the-art box-supervised methods on COCO \texttt{val2017} and \texttt{test-dev} split. The results of fully mask-supervised approaches are also listed as reference. Box2Mask-C achieves $32.2\%$ mask AP on COCO \texttt{val2017} split, which outperforms DiscoBox~\cite{iccv2021discobox} by 0.8\% AP ($32.2\%$ vs. $31.4\%$). Box2Mask-T further achieves $36.1\%$ mask AP with +$3.9\%$ improvement over Box2Mask-C. 
With ResNet-50 backbone, Box2Mask-C achieves $32.6\%$ mask AP on COCO \texttt{test2017} split, which outperforms the state-of-the-art method BoxInst~\cite{cvpr2021_boxinst} by $0.5\%$ AP ($32.6\%$ vs. $32.1\%$) and DiscoBox~\cite{iccv2021discobox} by 0.6\% AP ($32.6\%$ vs. $32.0\%$). In terms of high IoU threshold-based AP metric AP$_{75}$, Box2Mask-C surpasses BoxInst~\cite{cvpr2021_boxinst} and DiscoBox~\cite{iccv2021discobox} by $1.0\%$ and $0.8\%$, respectively. With the transformer-based framework, Box2Mask-T achieves a significant mask AP improvement (from $32.6\%$ to $36.7\%$), which is consistent with the results on COCO \texttt{val2017} split. 

\begin{figure*}[t]
	\centering
	\includegraphics[width=0.95\linewidth]{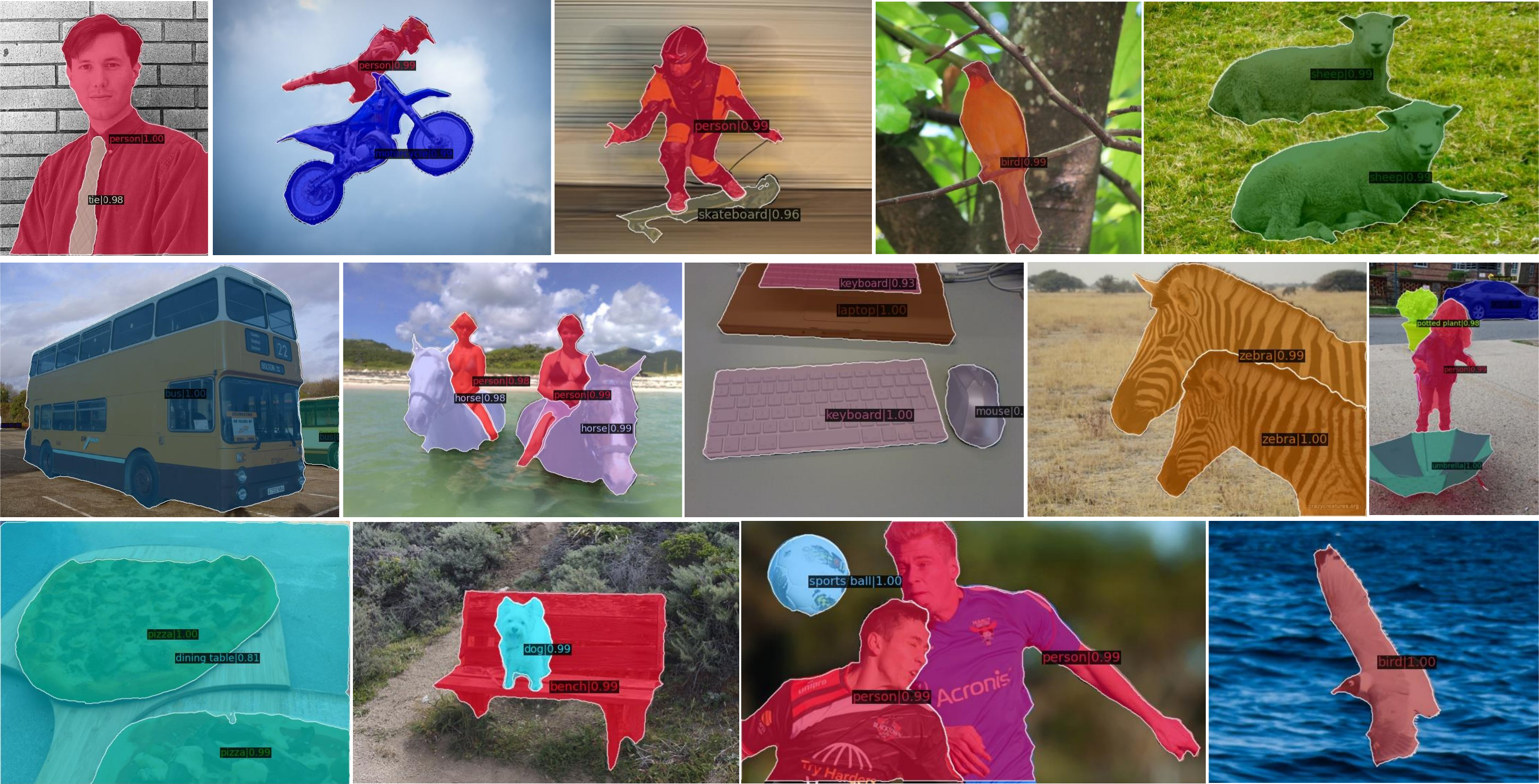}
	\caption{\textbf{Visualization of instance segmentation results} on COCO. The model is trained with only box annotations. Different colors encode different categories and instances. Best viewed on screen.}
	\label{fig:coco_vis_1}
\end{figure*}

With ResNet-101 backbone, Box2Mask-C outperforms  LIID~\cite{tpami2020leveraging} absolutely by $18.2\%$ mask AP ($34.2\%$ v.s. $16.0\%$). Though LIID only needs image-level supervision, it still requires the generic segment-based object proposals generated from MCG~\cite{tpmai2017mcg} during training. 
Box2Mask-C surpasses  A$^2$GNN~\cite{TPAMI2021affinity}, BBAM~\cite{cvpr2021bbam},  BoxCaseg~\cite{cvpr2021boxcaseg} and BoxInst~\cite{cvpr2021_boxinst} by $13.3\%$, $8.5\%$, $3.3\%$ and $1.0\%$ mask AP, respectively, by using the same ResNet-101 backbone. Box2Mask-C achieves 16.0\% AP$_S$ on small objects, which is slightly lower than BoxInst~\cite{cvpr2021_boxinst} by 0.2\%. This is because small objects lack rich features to distinguish from the background within the bounding box. However, Box2Mask-C obtains the best results for middle and large size objects, largely outperforming the second-best method BoxInst~\cite{cvpr2021_boxinst} by $2.4\%$ ($37.7\%$ vs. $35.3\%$) AP$_M$ and $3.2\%$ ($48.3\%$ vs. $45.1\%$) AP$_L$.

With the transformer-based framework, Box2Mask-T achieves $38.3\%$ mask AP, which is the best box-supervised instance segmentation result with the ResNet-101 backbone. It even performs better than some popular fully mask-supervised methods, such as Mask R-CNN~\cite{iccv2017maskrcnn} ($38.3\%$ vs. $35.7\%$),  SOLO~\cite{PMAI2021solo} ($38.3\%$ vs. $37.8\%$) and PolarMask~\cite{cvpr_2020polarmask} ($38.3\%$ vs. $32.1\%$). This indicates that our method narrows the performance gap between mask-supervised and box-supervised instance segmentation.
With Swin-Transformer base model (Swin-B) and Swin-Transformer large (Swin-L) model as the backbone, Box2Mask-T obtains 41.5\% and 42.4\% mask AP, respectively. It is on par with the recent fully-supervised approaches with ResNet-101, such as K-Net~\cite{nips2021knet},  QueryInst~\cite{ICCV2021_queryinst} and Mask2Former~\cite{cvpr2022mask2former}. 
Fig.~\ref{fig:coco_vis_1} visualizes the instance segmentation results of our method on COCO. 
Different categories are highlighted in different colors. One can see that our method accurately segments the boundaries of various objects in different scenes.

\subsubsection{Instance Segmentation on Aerial Images}
\textbf{Results on iSAID.} 
To explore the effectiveness of our method on aerial scene images, we conduct experiments on the challenging iSAID dataset. Table~\ref{tab:isaid} reports the mask AP results. 
We compare our method with both fully mask-supervised and weakly box-supervised approaches.
With ResNet-50 backbone, Box2Mask-C achieves $24.3\%$ AP, outperforming BoxInst~\cite{cvpr2021_boxinst} and DiscoBox~\cite{iccv2021discobox} by $6.8\%$ (from $17.5\%$ to $24.3\%$ ) and $2.9\%$ (from $21.4\%$ to $24.3\%$) under the same settings. This indicates that the pairwise affinity modeling methods are  susceptible to the label noises in complex aerial scenes, where the intra-class objects are densely-distributed and the inter-class instances are of high-similarity. 
With ResNet-101 as the backbone, Box2Mask-C obtains $25.8\%$ mask AP with 1$\times$ training schedule, outperforming BoxInst~\cite{cvpr2021_boxinst} and DiscoBox~\cite{iccv2021discobox} by $6.7\%$ and $3.2\%$ mask AP, respectively. By employing the longer 3$\times$ training scheme, our method achieves $26.6\%$ mask AP. It even outperforms the fully supervised method SOLO~\cite{wang2020solo} by $3.1\%$ AP, and performs on par with PolarMask~\cite{cvpr_2020polarmask} and SOLOv2~\cite{wang2020solov2} under the 1$\times$ training scheme. 

\begin{figure}[htbp]
\centering      
\includegraphics[width=0.55\linewidth]{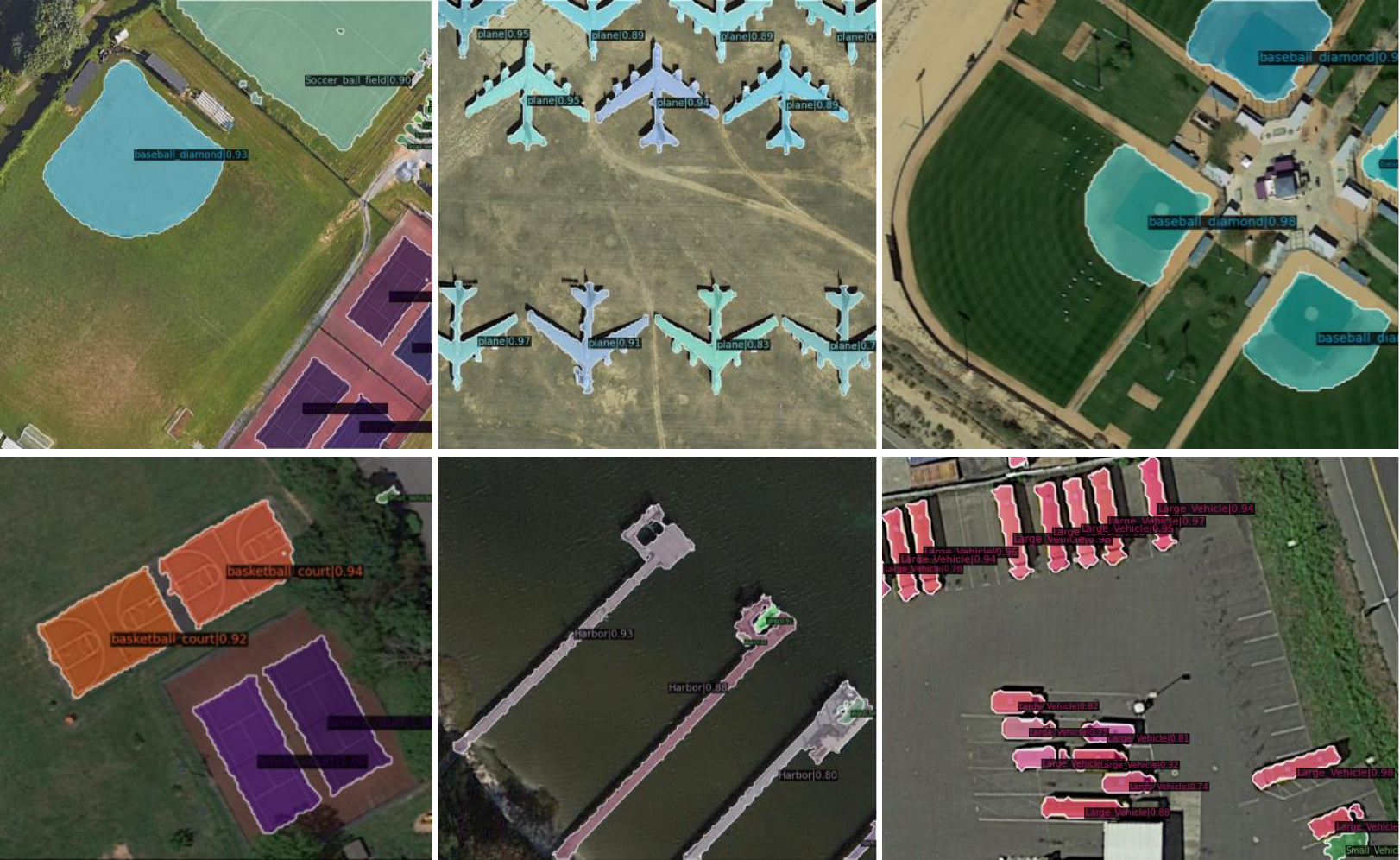}
\caption{\textbf{Visual examples of instance segmentation results} on iSAID \texttt{val}. The mask predictions are obtained on the high-resolution remote sensing images with only box supervision.}   
\label{fig:remote_sensing}   
\end{figure}

Our method cannot handle small objects very well, and it obtains $10.6\%$ AP$_S$ due to the large amount of small objects in remote sensing images. This is lower than the two-stage fully-supervised Mask R-CNN~\cite{iccv2017maskrcnn}. However, our method achieves comparable performance (AP$_L$) on large objects with Mask R-CNN~\cite{iccv2017maskrcnn} ($47.4\%$ vs. $47.9\%$).
We visualize some segmentation results on remote sensing scenes in Fig.~\ref{fig:remote_sensing}. One can see that our method still presents high quality segmentation results although the background is complicated and the instances of the same category are located close to each other. 
\begin{table}[t]
	\renewcommand\arraystretch{0.91}
	\centering
	\caption{Instance segmentation results on remote sensing image dataset iSAID \texttt{val}. The input size is 800$\times$800 for all models with 1$\times$ training schedule. ``${\ast}$" denotes the method with 3$\times$ training schedule.}
	\scalebox{0.93}{
	\setlength{\tabcolsep}{1.0mm}{
		\begin{tabular}{llcccccc}
			\toprule
			Method & Backbone & AP &AP$_{50}$ & AP$_{75}$ & AP$_{S}$ & AP$_{M}$ & AP$_{L}$ \\
			\midrule
			\multicolumn{8}{c}{\small \em mask-supervised} \\
			\midrule
			Mask R-CNN~\cite{iccv2017maskrcnn}  &ResNet-50  & 34.2 & 57.5 & 36.2& 19.6& 41.4 & 47.9\\
			PolarMask~\cite{cvpr_2020polarmask} & ResNet-50  &27.2 & 48.5 & 27.3 & - & -& -\\
			CondInst~\cite{eccv2020_condinst} & ResNet-50   & 31.8  & 56.4 & 31.5 & 15.3 & 41.0 & 50.2 \\
			SOLO~\cite{wang2020solo}  & ResNet-50  & 23.5 & 43.1 & 22.6 & 7.4 & 30.5 & 43.3   \\
			SOLOv2~\cite{wang2020solov2} & ResNet-50   & 28.9 & 50.6 & 28.8 & 11.6 & 37.8 & 48.3  \\
			SOLOv2$^{\ast}$~\cite{wang2020solov2} & ResNet-101    & 32.6 &54.4 &33.4 & 13.6 & 42.4 & 54.6    \\
			\midrule
			\multicolumn{8}{c}{\small \em box-supervised} \\
			\midrule
			BoxInst~\cite{cvpr2021_boxinst} & ResNet-50  & 17.5 & 42.9 & 11.4 & 8.8 & 23.9 & 37.2  \\
			BoxInst~\cite{cvpr2021_boxinst} & ResNet-101   & 19.1 & 45.2& 13.1 & 8.5 & 25.6 & 41.2  \\
			DiscoBox~\cite{iccv2021discobox} & ResNet-50      & 21.4 & 44.1 & 17.8 &  8.6 & 26.8 & 35.3  \\
			DiscoBox~\cite{iccv2021discobox} & ResNet-101    & 22.6& 45.3 & 19.4 & 9.2 & 28.6 & 38.7  \\
			\rowcolor{gray!10}
			Box2Mask-C & ResNet-50  & 24.3 & 48.1 & 20.7 & 9.9 & 29.9 & 40.9  \\
			\rowcolor{gray!10}
			Box2Mask-C & ResNet-101    &25.8 & 50.5 &22.4 & 10.6 &32.3 & 43.9  \\
			\rowcolor{gray!10}
			Box2Mask-C$^{\ast}$ & ResNet-101   & \textbf{26.6} & \textbf{50.6} & \textbf{23.8} &\textbf{10.6} & \textbf{33.6} &  \textbf{47.4} \\
			\bottomrule
	\end{tabular}}}
	\label{tab:isaid}
\end{table}

\textbf{Results on DOTA-v1.0.} 
Rotated object detection in aerial images is a challenging task, which has received increasing attention in recent years. 
Main-stream oriented detectors~\cite{ding2019learning, xu2020gliding, SASM_AAAI2022, li2022oriented, yang2022detecting} are mostly trained with accurate rotated bounding box (RBox) annotations with encoded orientation information on the interested objects, which costs intensive annotation labor$\footnote{The annotation price cost of the RBox is about 36.5\% higher than that of the HBox according to the data labeling platform at https://cloud.google.com/ai-platform/data-labeling/pricing.}$~\cite{yang2022h2rbox}.
Our method is trained with the ordinary horizontal bounding box (HBox) annotations. The rotated bounding box can be estimated by calculating the minimum compact surrounding rectangle from the mask predictions.
We conduct performance evaluation of our method on DOTA-v1.0~\cite{xia2018dota}, which is one of the largest datasets for rotated object detection. It contains 188,282 instance annotations on aerial scene images. Both RBox and HBox annotations are available. 

Table~\ref{tab:dota} shows the rotated object detection results. Using ResNet-50 as backbone, our method (Box2Mask-C) achieves $56.44\%$ and $60.16\%$ mAP with 1$\times$ and 3$\times$ training schedules, respectively. Besides, it achieves $63.35\%$ mAP by using ResNet-101 backbone. The performance is on par with the basic oriented detectors, including RepPoints-O~\cite{yang2019reppoints} and RetinaNet-O~\cite{lin2017focal}. Furthermore, our approach is able to provide the pose and shape information of rotated objects, not only the orientation in oriented detectors. 

\begin{table}[t]
	\renewcommand\arraystretch{0.86}
	\centering
	\caption{Performance of oriented object detection with the default AP$_{50}$(\%) on DOTA-v1.0 \texttt{test}. The ``RBox" denotes the supervision of fully rotated bounding box, and ``HBox" denotes weakly supervision by ordinary horizontal bounding box. ``-O'' denotes the oriented detector.}
	\scalebox{0.9}{
	\setlength{\tabcolsep}{3.8mm}{
		\begin{tabular}{llccc}
			\toprule
			Method & Backbone & Sched. &Size & mAP \\
			\midrule
			\multicolumn{5}{c}{\small \em fully RBox-supervised} \\
			\midrule
			RepPoints-O~\cite{yang2019reppoints}  & ResNet-50 & 1$\times$ &   1,024 & 59.44  \\
			RetinaNet-O~\cite{lin2017focal}  & ResNet-50  &  1$\times$ &   1,024 & 64.55  \\
			\midrule
			\multicolumn{5}{c}{\small \em HBox-supervised} \\
			\midrule
			BoxInst~\cite{cvpr2021_boxinst} & ResNet-50 & 1$\times$  & 960 &53.59  \\
			\rowcolor{gray!10}
			Box2Mask-C & ResNet-50 & 1$\times$  &960 & 56.44  \\
			\rowcolor{gray!10}
			Box2Mask-C & ResNet-50 &  3$\times$  & 960  & 60.16 \\
			\rowcolor{gray!10}
			Box2Mask-C  & ResNet-101 & 3$\times$ & 960  & 63.35 \\
			\bottomrule
	\end{tabular}}}
	\label{tab:dota}
\end{table}

\subsubsection{Instance Segmentation on Medical Images}
\begin{figure}[t]
\centering
\includegraphics[width=0.55\linewidth]{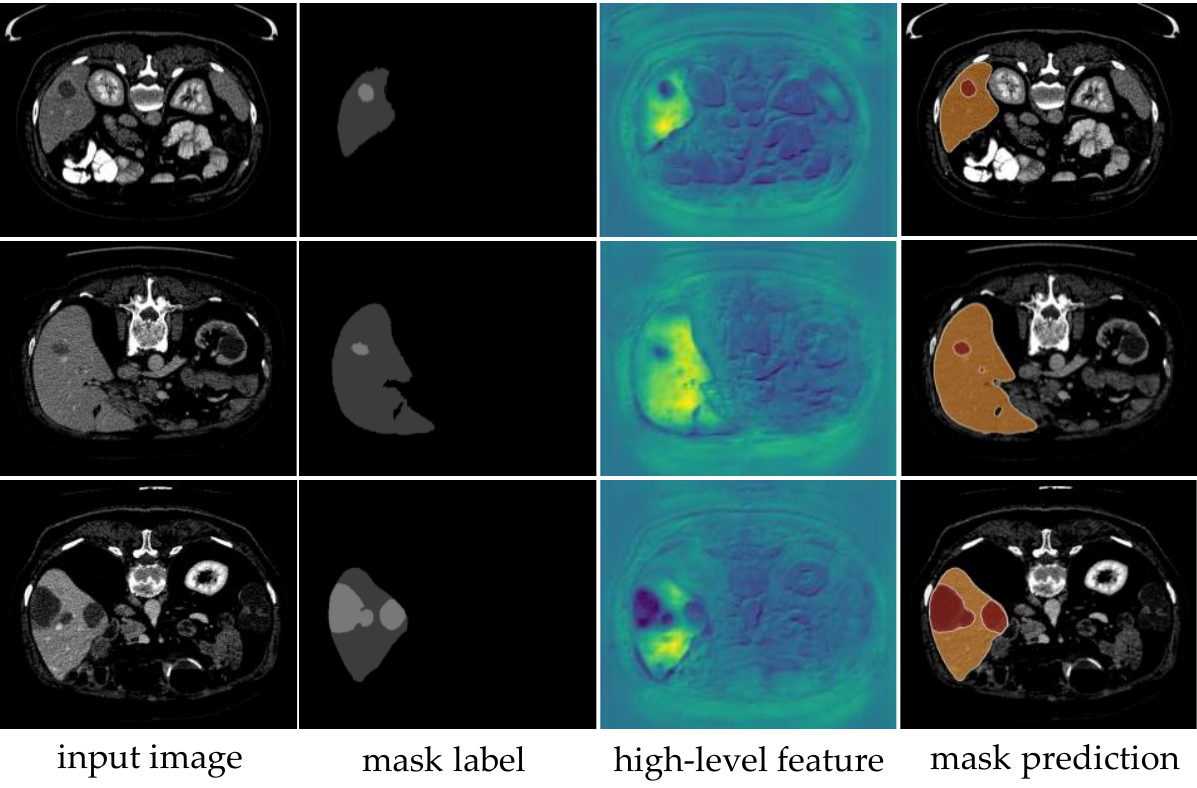}
\caption{\textbf{Visualization of instance segmentation} on LiTS \texttt{val}. The high-level deep semantic features are used as the input data for level-set evolution. Here the model is Box2Mask-T with ResNet-101.} 
\label{fig:lits_vis_1}
\end{figure}

In medical images, the background is highly similar to the foreground in appearance, making accurate instance segmentation very difficult. We validate the effectiveness of our method on the LiTS dataset. Table~\ref{tab:lits} shows the mask AP results. It can be seen that Box2Mask-C outperforms BoxInst~\cite{cvpr2021_boxinst} by $3.4\%$ mask AP ($52.3 \%$ vs. $48.9\%$) when ResNet-50 is used as the backbone. By using ResNet-101 as the backbone, Box2Mask-C achieves $52.8\%$ mask AP, which surpasses BoxInst~\cite{cvpr2021_boxinst} by $2.0\%$. 
Furthermore, Box2Mask-T gains $55.3\%$ mask AP with ResNet-101 backbone under the 2$\times$ training scheme.
Compared with the fully supervised instance segmentation approaches, Box2Mask-T obtains $80.0\%$ mask AP$_{50}$, which is close to SOLOv2~\cite{wang2020solov2} ($80.2\%$). Some visual results are shown in Fig.~\ref{fig:lits_vis_1}. One can see that although medical objects are highly similar to the background in appearance, our method can still evolve the level-set curves to accurately fit the object boundaries with the help of high-level features.

\begin{table}[t]
	\renewcommand\arraystretch{0.88}
	\centering
	\caption{Instance segmentation results on medical image dataset LiTS \texttt{val}. Unless specified, all models are trained with 640$\times$640 input size with 1$\times$ training schedule. ``${\ast}$" denotes the method using multi-scale training with 2$\times$ training schedule.} 
	\scalebox{0.88}{
    \setlength{\tabcolsep}{1.0mm}{
	\begin{tabular}{ llccccc}
		\toprule
		Method  & Backbone & AP &AP$_{50}$ & AP$_{75}$ & Liver & Tumor \\
		\midrule
		Mask R-CNN~\cite{iccv2017maskrcnn}   & ResNet-50 & 63.8 & 83.0 & 69.9 & 85.7 & 41.9\\
		SOLOv2~\cite{wang2020solov2}   & ResNet-50 & 63.1 & 80.2 & 66.6 & 89.3 & 36.9  \\
		\midrule
		BoxInst~\cite{cvpr2021_boxinst}  & ResNet-50  & 48.9 & 77.3 & 49.8  & 72.9 & 25.0 \\
		BoxInst~\cite{cvpr2021_boxinst}   & ResNet-101 & 50.8 & 79.8 & 52.2 & 73.9 & 27.5\\
		\rowcolor{gray!10}
		Box2Mask-C   & ResNet-50  & 52.3 & 79.3 & 55.4  & 72.9 & 31.8    \\
		\rowcolor{gray!10}
		Box2Mask-C   & ResNet-101 & 52.8 & 79.5 & 56.0 & 73.8 & 31.7  \\
		\rowcolor{gray!10}
		\rowcolor{gray!10}
		Box2Mask-T$^{\ast}$  & ResNet-50 & 54.9 & 80.0 & 57.6 &  77.6 & 32.2  \\
		\rowcolor{gray!10}
		Box2Mask-T$^{\ast}$  & ResNet-101 & \textbf{55.3} & \textbf{80.0} & \textbf{58.4} & \textbf{78.0} & \textbf{32.5}  \\
		\bottomrule
\end{tabular}}}
\label{tab:lits}
\end{table}

\begin{table}[t]
\renewcommand\arraystretch{0.90}
\centering
\caption{Object detection performance on the ICDAR2019 ReCTS \texttt{test} dataset. All methods are supervised by horizontal bounding box. The predicted bounding box are generated by the mask prediction.}
\scalebox{0.88}{
\setlength{\tabcolsep}{4.0mm}{
	\begin{tabular}{llccc}
		\toprule
		Method & Backbone & AP & AP$_{50}$ & AP$_{75}$\\
		\midrule
		DiscoBox~\cite{iccv2021discobox}  & ResNet-50 & 41.6 & 58.8 & 46.5 \\
		BoxInst~\cite{cvpr2021_boxinst}  & ResNet-50 & 42.6  &  74.1 & 44.5 \\
		\rowcolor{gray!10}
		Box2Mask-C  & ResNet-50 &  44.6 & 64.0& 50.4 \\
		\rowcolor{gray!10}
		Box2Mask-T  & ResNet-101 & \textbf{53.9} & \textbf{78.9} & \textbf{59.8}  \\
		\bottomrule
\end{tabular}}}
\label{tab:icdar}
\end{table}

\begin{figure}[t]
	\centering   
	\includegraphics[scale=0.65]{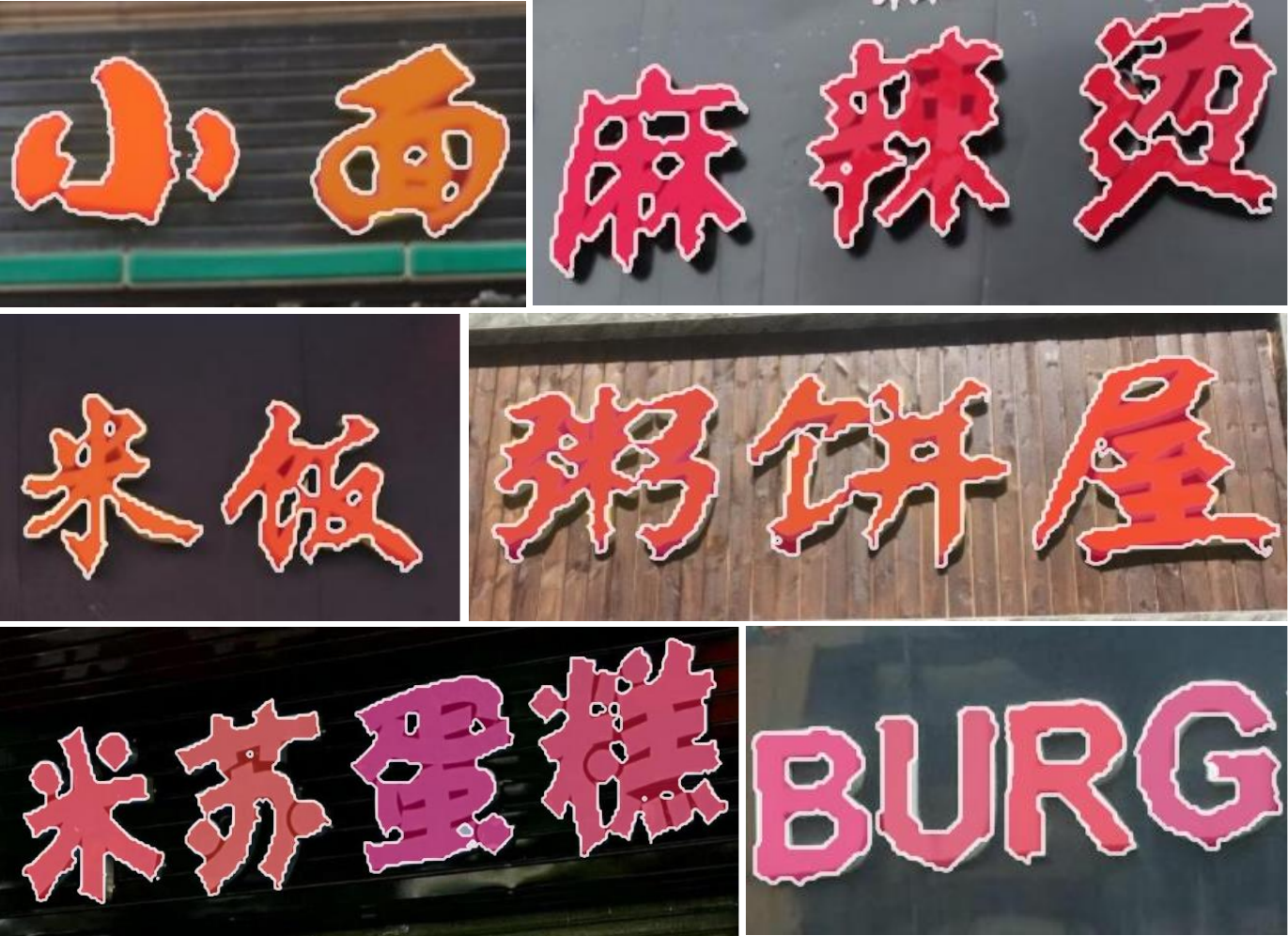}
	\caption{\textbf{Visualization of character instance segmentation results} on ICDAR 2019 ReCTS. The model is trained with character-wise box annotations.}   
	\label{fig:icdar}   
\end{figure}

\subsubsection{Character Instance Segmentation}

To demonstrate the generalization capability of our approach, we conduct experiments to estimate the character masks with character-wise box annotations on ICDAR 2019 ReCTS~\cite{zhang2019icdar}. 
Since the mask annotations are unavailable, it is impossible to report the mask AP. Instead, we report the object detection performance. The predicted bounding boxes are calculated from the extreme (top, bottom, left and right) points of the predicted masks.
As shown in Table~\ref{tab:icdar},  Box2Mask-C obtains $44.6\%$ box AP, which outperforms DiscoBox~\cite{iccv2021discobox} and BoxInst~\cite{cvpr2021_boxinst} by $3.0\%$ AP and $2.0\%$ AP, respectively. Box2Mask-T achieves the best performance with $53.9\%$ AP, which outperforms Box2Mask-C by a large margin. 
We visualize some character-wise instance segmentation results in Fig.~\ref{fig:icdar}.
The character-wise text masks could provide useful cues for detecting and recognizing text of arbitrary shapes. We believe that the capability of our model in predicting character masks may inspire new solutions for the tasks of text detection and recognition.

\begin{table}[t]
	\renewcommand\arraystretch{0.90}
	\centering
	\caption{Deep variational instance segmentation results on COCO \texttt{val}. ``Sup." denotes the form of supervision, i.e., Mask ($\mathcal{M}$) or Box ($\mathcal{B}$) (the same in other tables). 
		Our method is supervised by only box annotations but it achieves highly competitive results. Box2Mask-T achieves the best $37.5\%$ AP.}
	\scalebox{0.88}{
	\setlength{\tabcolsep}{1.5mm}{
		\begin{tabular}{llcccc}
			\toprule
			Method & Backbone & Sup. & AP & AP$_{50}$ & AP$_{75}$  \\
			\midrule
			DeepSnake~\cite{peng2020deepsnake}  &DLA-34~\cite{yu2018dla34} & $\mathcal{M}$ &30.5 & - & -\\
			Levelset R-CNN~\cite{eccv2020levelset} &ResNet-50 & $\mathcal{M}$ & 34.3 & - & - \\
			DVIS-700\cite{yuan2020deep} & ResNet-50 & $\mathcal{M}$ &32.6 & 53.4 & 35.0 \\
			DVIS-700\cite{yuan2020deep} & ResNet-101 & $\mathcal{M}$ & 35.7 & 58.0 & 37.5  \\
			\midrule
			\rowcolor{gray!10}
			Box2Mask-C & ResNet-50 & $\mathcal{B}$ & 32.2 & 54.4 & 32.8  \\
			\rowcolor{gray!10}
			Box2Mask-C & ResNet-101 & $\mathcal{B}$ & 34.0 & 56.9 &34.8  \\
			\rowcolor{gray!10}
			Box2Mask-T & ResNet-50 & $\mathcal{B}$ & 36.1 & 60.8 & 36.4 \\
			\rowcolor{gray!10}
			Box2Mask-T & ResNet-101 & $\mathcal{B}$ & \textbf{37.5} & \textbf{63.7} & \textbf{38.0}  \\
			\bottomrule
	\end{tabular}}}
	\label{tab:deepvaria}
\end{table}
\begin{table}[t]
	\renewcommand\arraystretch{0.90}
	\centering
	\caption{The mask AP and inference speed of our method. Typical mask-supervised and box-supervised methods are listed here. The inference time is measured by using a single V100 GPU.}
	\scalebox{0.88}{
	\setlength{\tabcolsep}{3.5mm}{
		\begin{tabular}{llccc}
			\toprule
			Method & Backbone & Sup. & FPS & AP   \\
			\midrule
			Mask R-CNN~\cite{iccv2017maskrcnn}  & ResNet-50 & $\mathcal{M}$ & 19.4 & 34.7  \\
			SOLO~\cite{wang2020solo}  & ResNet-50 & $\mathcal{M}$ & 14.2 &  36.8   \\
			SOLOv2~\cite{wang2020solov2} & ResNet-50 & $\mathcal{M}$ & 19.6 & 38.2  \\
			CondInst~\cite{eccv2020_condinst} & ResNet-50 & $\mathcal{M}$ & 16.0  & 37.8  \\
			\midrule
			DiscoBox~\cite{iccv2021discobox} & ResNet-50  &$\mathcal{B}$ & \textbf{16.8} &32.0   \\
			BoxInst~\cite{cvpr2021_boxinst} & ResNet-50  &$\mathcal{B}$ & 16.0  &32.1  \\
			\rowcolor{gray!10}
			Box2Mask-C & ResNet-50 & $\mathcal{B}$ & 11.5 & 32.6   \\
			\rowcolor{gray!10}
			Box2Mask-T & ResNet-50 & $\mathcal{B}$ & 8.7 & 36.7 \\
			\rowcolor{gray!10}
			Box2Mask-T & ResNet-101 & $\mathcal{B}$ & 7.9 & \textbf{38.3}\\
			\bottomrule
	\end{tabular}}}
	\label{tab:speed}
\end{table}

\subsection{Deep Variational-based Instance Segmentation}
To further evaluate the proposed level-set based approach, we compare it with other deep variational-based instance segmentation approaches, including DeepSnake~\cite{peng2020deepsnake}, 
Levelset R-CNN~\cite{eccv2020levelset} and DVIS-700~\cite{yuan2020deep}. Among them, 
DeepSnake~\cite{peng2020deepsnake} is based on the classical snake method~\cite{kass1988snakes}. 
Levelset R-CNN~\cite{eccv2020levelset} and DVIS-700~\cite{yuan2020deep} are built upon level-set evolution. Please note these three compared methods are all fully supervised by the mask annotations. 

As shown in Table~\ref{tab:deepvaria}, Box2Mask-C achieves comparable results to the fully supervised variational-based methods, even outperforming  DeepSnake~\cite{peng2020deepsnake} by $0.7\%$ AP on COCO \texttt{val} dataset using ResNet-50 as backbone. Box2Mask-T surpasses Levelset R-CNN~\cite{eccv2020levelset} and DVIS-700~\cite{yuan2020deep} by $1.8\%$ AP ($36.1\%$ vs. $34.3$)  and  $3.5\%$ AP ($36.1\%$ vs. $32.6$)  with ResNet-50 backbone, respectively. Using ResNet-101, Box2Mask-T achieves the best result of $37.5\%$ AP, outperforms all these fully supervised deep variational-based methods.

\subsection{Inference Speed}
The inference speed and accuracy of typical mask-supervised and box-supervised methods are reported in Table~\ref{tab:speed}. 
The Box2Mask-C with ResNet-50 backbone achieves $11.5$ FPS and $32.6\%$ mask AP. The speed of our method is slower than DiscoBox~\cite{iccv2021discobox} and BoxInst~\cite{cvpr2021_boxinst} but with better accuracy. 
Compared with Box2Mask-C, Box2Mask-T needs more computational cost. Box2Mask-T with ResNet-50 backbone runs at a speed of $8.7$ FPS but brings a significant accuracy improvement to $36.7\%$ mask AP. 
By using a stronger backbone ResNet-101, Box2Mask-T can run at a similar FPS of $7.9$ but achieves the best box-supervised performance of $38.3\%$ mask AP.

\subsection{Ablation Experiments}
We conduct ablation studies on Pascal VOC dataset to examine the effectiveness of each module in our proposed Box2Mask framework.

\textbf{Level-set Energy.} We firstly investigate the impact of level-set energy functional with different settings in Box2Mask-C. Table~\ref{tab:lossterm} shows the evaluation results. Our method achieves $27.1\%$ AP by using the box projection function as $\mathcal{F}_{\phi_0}$ to initialize the boundary during training. This demonstrates that the initialization for level-set function is effective to generate the initial boundary. When the original image $I_{u}$ is employed as the input data term in Eq.~\ref{eq8}, our method achieves $30.6\%$ AP. When the deep features $I_{f}$ are employed as the extra input data, our method achieves a better performance of $33.8\%$ AP. This demonstrates that both original image and high-level features can provide useful information for robust level-set evolution. Note that the above results are obtained by constraining curve evolution within the bounding box $\mathcal{B}$ region. When the global region with the full-image size is regarded as the $\Omega$, there is a noticeable performance drop ($30.6\%$ vs. $28.3\%$ and $33.8\%$ vs. $28.8\%$) with the same input data terms. 

\begin{table}[ht]
	\renewcommand\arraystretch{0.92}
	\centering
	\caption{The impact of \textbf{level-set energy} with different settings. $I_{u}$ and $I_{f}$ denote that the low-level image feature and high-level deep feature are used as the input data term, respectively. $\mathcal{B}$ and $I$ represent the $\Omega$ space of bounding box or the full-image region for level-set evolution.}
	\scalebox{0.88}{
	\setlength{\tabcolsep}{1.6mm}{
			\begin{tabular}{cccccccc}
				\toprule
				$\mathcal{F}_{\phi_0}$ & $\mathcal{F_{\phi}}({I_{u}})$ & $\mathcal{F}_{\phi}{(I_{f})}$ & $\Omega \in \mathcal{B}$ & $\Omega \in I$ &  AP &AP$_{50}$ & AP$_{75}$    \\
				\midrule
				\cmark &  & & \cmark & & 27.1 & 60.3 & 21.1    \\ 
				\cmark &\cmark & & \cmark& & 30.6 & 62.0 & 27.6   \\
				\cmark &\cmark & & &\cmark & 28.3 & 60.4 & 23.8   \\
				\cmark &\cmark &\cmark &\cmark & & \textbf{33.8} & \textbf{62.5} & \textbf{32.9}    \\
				\cmark &\cmark &\cmark & &\cmark & 28.8 & 57.2 & 26.5    \\
				\bottomrule
		\end{tabular}}}
		\label{tab:lossterm}
	\end{table}

\begin{table}[ht]
	\renewcommand\arraystretch{0.90}
	\centering
	\caption{The effectiveness of \textbf{deep structural features} with different input guidance of tree filter~\cite{nips2019learnable}.} 
	\scalebox{0.88}{
	\setlength{\tabcolsep}{1.8mm}{
		\begin{tabular}{ lcccccc}
			\toprule
			\textit{guidance} &  AP &AP$_{50}$ & AP$_{75}$ & AP$_{S}$ & AP$_{M}$ & AP$_{L}$ \\
			\midrule
			~~~~\xmark~  &   30.8  & 62.2 & 27.9 & 6.1 & 24.2 & 40.7   \\
			low-level   & 28.2   & 51.5 & 27.6 & \textbf{7.7} & 23.4 & 35.6  \\
			high-level  &  30.8 & 62.1 & 27.5 & 6.8 & 23.9 & 41.2   \\
			low \& high-level &  \textbf{33.8} & \textbf{62.5} & \textbf{32.9} & 6.7 & \textbf{26.8} & \textbf{45.0}    \\
			\bottomrule
	\end{tabular}}}
	\label{tab:treefilter}
\end{table}
\textbf{Deep Structural Feature with Tree Filters.} 
We study the impact of tree filters~\cite{nips2019learnable}, which can capture long-range feature dependency, on generating  structural features as the input to drive level-set evolution. Table~\ref{tab:treefilter} shows the results. 
Without using the tree filters, our method achieves $30.8\%$ mask AP. By using only the low-level image feature as the guidance of tree filters, the performance will drop from $30.8\%$ to $28.2\%$. This is because the low-level features can be noisy for the tree filters to build structural features. By using the low-level and high-level features together as the guidance, the tree filters can bring $3.0\%$ AP improvement, achieving $33.8\%$ mask AP.

\textbf{Local Consistency Module.} 
We study the effectiveness of the proposed local consistency module (LCM), which aims to maintain the local affinity consistency and avoid the intensity inhomogeneity in level-set evolution.  Table~\ref{tab:LAC} shows the results. Without using LCM for level-set prediction, our approach can obtain $33.8\%$ AP. In LCM, different dilation rates can be used to control the scope of local region. One can see that by applying LCM with dilation rate 3, $36.3\%$ AP can be achieved, with $+2.5\%$ AP improvement. Actually, dilation rates 1, 2 and 3 lead to similar performance. When the local region is enlarged too much by using dilation rate 4 or 5, the performance drops. This demonstrates that LCM benefits local affinity consistency, instead of global dependency.

\begin{table}[t]
	\renewcommand\arraystretch{0.88}
	\centering
	\caption{The effectiveness of \textbf{local consistency module (LCM)}  using different dilation rates.} 
	\scalebox{0.88}{
	\setlength{\tabcolsep}{2.9mm}{
			\begin{tabular}{ccccccc}
				\toprule
				\textit{dilation}  &  AP &AP$_{50}$ & AP$_{75}$ & AP$_{S}$ & AP$_{M}$ & AP$_{L}$ \\
				\midrule
				\xmark & 33.8 & 62.5 & 32.9 &6.7 & 26.8 & 45.0 \\
				1 & 36.1 & \textbf{64.6} & 35.1 & 10.2 & 29.0 & 47.2\\
				2 & 36.2 & 64.0 & \textbf{36.0} & 8.9 & 28.9 & 47.4 \\
				3 & \textbf{36.3} & 64.2 & 35.8 & \textbf{9.5}& \textbf{29.3} & \textbf{47.6} \\
				4 & 35.6 & 63.0 &35.9 & 8.7 & 29.6 & 46.5 \\
				5 & 33.5 & 59.1 & 34.2& 5.4 & 25.7 & 45.5 \\
				\bottomrule
		\end{tabular}}}
		\label{tab:LAC}
\end{table}

\textbf{Training Schedule.} We evaluate the proposed Box2Mask-C and Box2Mask-T models using different training schedules. Table~\ref{tab:trainingschdule} reports the results of Box2Mask-C using 12 epochs (1$\times$) with single-scale input size, as well as 36 epochs (3$\times$) with multi-scale input size. 
For Box2Mask-T, we also evaluate its performance using 30 epochs and 50 epochs training schedules.
It can be observed that a longer training schedule can benefit our proposed models, bringing visible AP improvement ($36.3\%$ vs. $38.0\%$, $39.4\%$ vs. $41.4\%$).  

\begin{table}[t]
	\renewcommand\arraystretch{0.86}
	\centering
	\caption{\textbf{Training schedules} for Box2Mask-C and Box2Mask-T models. ``MS" denotes multi-scale training. ``30e" and ``50e" denote 30-epoch and 50-epoch training schedules.} 
	\scalebox{0.88}{
	\setlength{\tabcolsep}{2.9mm}{
		\begin{tabular}{ cccccc}
			\toprule
			\textit{model} & \textit{sched.} & MS &  AP &AP$_{50}$ & AP$_{75}$  \\
			\midrule
			\multirow{2}*{Box2Mask-C} & 1$\times$  &  & 36.3 & 64.2 & 35.8   \\
			~ & 3$\times$  & \cmark  &38.0 & 65.9 & 38.2   \\ 
			\midrule
			\multirow{2}*{Box2Mask-T} & 30e & \cmark & 39.4 & 67.6 & 39.3  \\
			& 50e & \cmark & \textbf{41.4} & \textbf{68.9} & \textbf{42.1} \\
			\bottomrule
	\end{tabular}}}
	\label{tab:trainingschdule}
\end{table}

\begin{table}[t]
	\renewcommand\arraystretch{0.86}
	\centering
	\caption{The impact of \textbf{balance weights $\beta{}_1$ and $\beta{}_2$ in matching cost}  for the box-level matching assignment in Box2Mask-T.} 
	\scalebox{0.88}{
		\setlength{\tabcolsep}{2.5mm}{
			\begin{tabular}{cccccccc}
				\toprule
				$\beta{}_1$ & $\beta{}_2$ &  AP &AP$_{50}$ & AP$_{75}$ & AP$_{S}$ & AP$_{M}$ & AP$_{L}$ \\
				\midrule
				2.0 & 3.0 & 39.0 &  67.6 & 38.8 &7.0 & \textbf{27.8} & 53.3  \\
				2.0 & 4.0 & 39.1  & 66.7  & 38.8 & \textbf{7.7} &  27.4 & 53.7\\
				2.0 & 6.0 &  \textbf{39.4} & \textbf{67.6}  & \textbf{39.3} & 6.8 & 27.4 & \textbf{53.9}  \\ 
				3.0 & 6.0 &  39.1 &  67.3 &  39.2 & 6.5 & 27.2 & 53.8  \\
				1.0 & 6.0 &  38.1 &  66.1  & 38.0 & 7.3 & 26.6 & 51.8  \\
				\bottomrule
	\end{tabular}}}
	\label{tab:matchingweight}
\end{table}
\begin{table}[t]
	\renewcommand\arraystretch{0.86}
	\centering
	\caption{The effectiveness of the \textbf{number of MSDeformAtten} layers in the pixel decoder of Box2Mask-T. }
	\scalebox{0.88}{ 
		\setlength{\tabcolsep}{2.6mm}{
			\begin{tabular}{ccccccc}
				\toprule
				\textit{number} &  AP &AP$_{50}$ & AP$_{75}$ & AP$_{S}$ & AP$_{M}$ & AP$_{L}$ \\
				\midrule
				1 &  36.8 &  65.7 & 36.1 & 6.1 & 26.8 & 50.0  \\
				2 &  38.9 & 67.7 & 38.7 & 6.9 & 27.8 & 52.8 \\
				4  & 39.3 & 67.1 & 38.9 & 7.2 & 28.0 & 53.6 \\
				5  & \textbf{39.4} & \textbf{67.9} & \textbf{39.7} & \textbf{7.8} & \textbf{27.9} & \textbf{53.8} \\
				6  & 39.3& 67.3 & 39.7 &6.8 &27.6 &  53.4 \\
				\bottomrule
	\end{tabular}}}
	\label{tab:msdeformatten}
\end{table}

\textbf{Balance Weights of Matching Cost.} 
We explore the setting of balance weights in Eq.~\ref{mathing_cost_eq} for box-level matching assignment in transformer-based framework. The hyper-parameters
$\beta  {}_1$ and $\beta  {}_2$ denote the importance of category and segmentation in matching cost, respectively. Table~\ref{tab:matchingweight} reports the experimental results. It can be seen that the best performance ($39.4\%$ AP) is obtained when we set $\beta_1 = 2.0$ and $\beta_1 = 6.0$.

\textbf{The Number of MSDeformAtten Layers.}  We further study the selection of the number of multi-scale deformable
attention (MSDeformAtten) layers in pixel decoder to encode the robust mask features.
The results are reported in Table~\ref{tab:msdeformatten}.
When the number of MSDeformAttn is 1, we achieve 36.8$\%$ mask AP. When two MSDeformAtten layers are used, a visible improvement of +$2.1\%$ is obtained ($38.9\%$ vs. $36.8\%$). As the number is 5, the best performance $39.4\%$ is achieved on Pascal VOC.

\section{Conclusion}
This paper presented a highly effective box-supervised instance segmentation approach, namely Box2Mask, by iteratively learning a series of level-set functions to obtain accurate segmentation mask predictions. 
The CNN-based and transformer-based architectures of Box2Mask were respectively developed to perform the level-set evolution in a single-stage framework.
Both the input image its deep high-level features were employed as the inputs to the network for evolving the level-set curves, where a box projection function was employed to initialize the object boundary. By minimizing the fully differentiable energy function, the level-set curve for each instance was iteratively optimized within its corresponding bounding box annotation. We conducted extensive experiments on five benchmarks covering different types of images. The proposed Box2Mask method recorded new state-of-the-arts on all the datasets, clearly demonstrating its effectiveness and generality for instance segmentation. Our work significantly narrowed the performance gap between the fully mask-supervised and box-supervised approach, making instance segmentation with simple bounding box annotations more practical.

{\small
	\bibliographystyle{ieee_fullname}
	\bibliography{mainbib}
}

\end{document}